\documentclass[10pt,twocolumn,letterpaper]{article}

\usepackage{cvpr}              

\definecolor{cvprblue}{rgb}{0.21,0.49,0.74}
\definecolor{lightgray}{rgb}{0.93, 0.93, 0.93}
\usepackage[pagebackref,breaklinks,colorlinks,allcolors=cvprblue]{hyperref}
\usepackage{multirow}
\usepackage{algorithm}
\usepackage{algorithmic}
\usepackage{amsmath}
\usepackage{amssymb}
\usepackage[table]{xcolor} 

\def\paperID{} 
\def\confName{CVPR}
\def\confYear{2026}

\title{Learning Generalizable 3D Medical Image Representations from Mask-Guided Self-Supervision}

\author{
Yunhe Gao,\quad Yabin Zhang,\quad Chong Wang,\quad Jiaming Liu,\quad Maya Varma, \\
Jean-Benoit Delbrouck,\quad Akshay Chaudhari,\quad Curtis Langlotz\vspace{+0.5em}\\
Stanford University
}

\begin{document}
\maketitle
\begin{abstract}

Foundation models have transformed vision and language by learning general-purpose representations from large-scale unlabeled data, yet 3D medical imaging lacks analogous approaches. Existing self-supervised methods rely on low-level reconstruction or contrastive objectives that fail to capture the anatomical semantics critical for medical image analysis, limiting transfer to downstream tasks. We present MASS (MAsk-guided Self-Supervised learning), which treats in-context segmentation as the pretext task for learning general-purpose medical imaging representations. MASS's key insight is that automatically generated class-agnostic masks provide sufficient structural supervision for learning semantically rich representations. By training on thousands of diverse mask proposals spanning anatomical structures and pathological findings, MASS learns what semantically defines medical structures: the holistic combination of appearance, shape, spatial context, and anatomical relationships. We demonstrate effectiveness across data regimes: from small-scale pretraining on individual datasets (20-200 scans) to large-scale multi-modal pretraining on 5K CT, MRI, and PET volumes, all without annotations. MASS demonstrates: (i) few-shot segmentation on novel structures, (ii) matching full supervision with only 20-40\% labeled data while outperforming self-supervised baselines by over 20 in Dice score in low-data regimes, and (iii) frozen-encoder classification on unseen pathologies that matches full supervised training with thousands of samples. Mask-guided self-supervised pretraining captures broadly generalizable knowledge, opening a path toward 3D medical imaging foundation models without expert annotations. Code is available \href{https://github.com/Stanford-AIMI/MASS}{here}.

\end{abstract}    
\section{Introduction}
\label{sec:intro}


Large-scale self-supervised learning (SSL) has transformed artificial intelligence. Foundation models like GPT~\cite{brown2020language}, CLIP~\cite{radford2021learning}, and DINO~\cite{caron2021emerging,oquab2023dinov2,simeoni2025dinov3} learn general-purpose representations from vast unlabeled data that transfer broadly with minimal adaptation. For example, linear probes on frozen features often rival specialized models~\cite{assran2023self}, and few-shot learning enables rapid deployment to new domains~\cite{zhou2022conditional,ayzenberg2024dinov2,bou2024exploring}. A single pretrained model serves diverse downstream applications, from classification to segmentation~\cite{kang2023distilling}. However, 3D medical imaging lacks an equivalent foundation.

\begin{figure*}[t]
    \centering
    \includegraphics[width=0.85\linewidth]{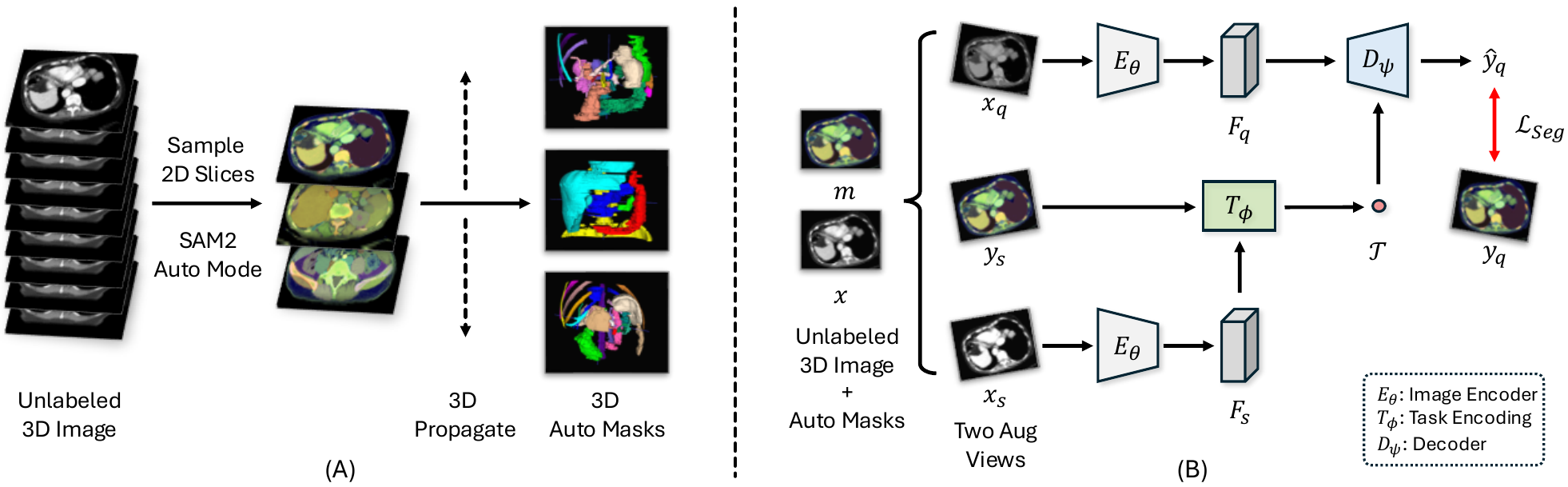}
    \caption{MASS framework overview. (A) Annotation-free mask generation: SAM2 generates class-agnostic masks by sampling 2D slices from unlabeled 3D images, applying automatic segmentation, and propagating masks through volumes. B) Mask-guided self-supervised learning: For each training step, we sample an image $x$ and its auto-generated masks $m$. We then create two augmented views: a \emph{reference} $(x_s, y_s)$ and a \emph{query} $(x_q, y_q)$. The model extracts a task embedding from the reference mask and uses it to predict the corresponding region in the query view. By solving many such in-context segmentation tasks, the model learns generalizable, semantically rich representations.}
    \label{fig:framework}
    \vspace{-1em}
\end{figure*}

Medical imaging presents unique challenges that expose the limitations of existing pretraining paradigms. Unlike natural images where semantically meaningful content occupies sparse regions against extensive backgrounds, most voxels in a medical scan carry clinical significance~\cite{van2020radiomics}: bone density reveals fractures, soft tissue textures distinguish tumors from healthy parenchyma, and vascular patterns diagnose ischemia. Moreover, spatial precision is clinically critical: a 2 cm tumor needs different treatment than a 5 cm mass, while hemorrhage location determines etiology and prognosis. General-domain self-supervised methods optimize for either global image-level features~\cite{chen2020simple,caron2021emerging} or low-level texture reconstruction~\cite{he2022masked}, failing to capture these local, spatially-grounded semantics. Supervised pretraining approaches~\cite{liu2023clip,li2025well,gao2024training}, while effective within their annotation scope, remain constrained by predefined class taxonomies and cannot scale to the thousands of anatomical variants and pathologies encountered across clinical practice. What medical imaging needs is a pretraining objective that learns dense, spatially-precise representations across the full spectrum of anatomical and pathological concepts without exhaustive human annotation.

In this work, we introduce a novel self-supervised learning framework called \textit{MA}sk-guided \textit{S}elf-\textit{S}upervised learning (MASS). The core intuition is simple: semantic segmentation is the most clinically aligned pretext task. Clinicians reason by identifying what a structure is and where it is located (tumors, organs, vessels, lesions). This makes semantic segmentation a natural candidate for in-context learning that is both semantically meaningful and spatially precise. The traditional barrier is annotation cost: pixel-level delineation of 3D medical scans requires expertise and substantial expense. MASS circumvents this through a simple yet powerful idea: learn strong semantic representations from weak, class-agnostic masks. Modern segmentation tools~\cite{achanta2012slic,kirillov2023segment,ravi2024sam} readily generate thousands of region proposal masks per volume based on intensity boundaries and morphological coherence. While these masks lack semantic labels and contain noise, they capture anatomically and pathologically meaningful regions spanning healthy organs, vascular networks, tumors, lesions, and other clinical findings. MASS formulates pretraining as learning to perform in-context segmentation~\cite{butoi2023universeg,gao2025show} across diverse mask proposals: given reference examples showing a region of interest, predict the same structure in query views. This objective forces the model to learn what semantically defines medical structures beyond superficial boundaries, encoding the holistic combination of appearance, shape, spatial context, and anatomical relationships. Critically, diverse weak supervision suffices: the model learns robust, transferable representations by solving thousands of approximately correct segmentation tasks. These learned features prove effective not only for dense prediction tasks but also for other downstream applications, like global classification, demonstrating that mask-guided pretraining captures broadly generalizable medical imaging knowledge.

MASS is simple, scalable, and effective. Our main contributions are:

\begin{itemize}
\item \textit{A novel self-supervised framework for 3D medical imaging.} We present MASS, which uses in-context segmentation as the pretext task and automatically generated class-agnostic masks as structural anchors, eliminating the need for expert annotations while learning semantically rich and generalizable representations.

\item \textit{Annotation-free scalability across data regimes.} MASS performs effectively from small-scale pretraining on individual datasets (20-200 scans) to large-scale multi-modal pretraining on 5K multi-modal CT, MRI, and PET volumes, demonstrating consistent improvements as data scale and diversity increase.

\item \textit{Strong few-shot capabilities.} Pretrained MASS models directly learn medical concepts and can perform few-shot in-context segmentation without finetuning. With finetuning on just 20-40\% of training data, MASS matches full supervision performance and substantially outperforms all prior self-supervised methods, with improvements exceeding 20 Dice points in extreme low-data settings.

\item \textit{Broad generalization to unseen tasks.} On frozen-encoder classification of unseen pathologies, MASS matches or exceeds full supervised training with thousands of labeled samples, demonstrating that mask-guided pretraining captures broadly transferable medical imaging knowledge beyond segmentation.
\end{itemize}

\section{Related Work}
\label{sec:related_work}

\textbf{Self-Supervised Learning.}
Self-supervised learning (SSL) has achieved remarkable success in computer vision through various paradigms. Contrastive methods like MoCo~\cite{he2020momentum} and SimCLR~\cite{chen2020simple} learn representations by maximizing agreement between different augmented views of the same image. Masked image modeling approaches such as MAE~\cite{he2022masked} and SimMIM~\cite{xie2022simmim} predict masked patches to learn visual features. Self-distillation methods including DINO~\cite{caron2021emerging} and DINOv2~\cite{oquab2023dinov2} train student networks to match teacher predictions without labels. These methods learn powerful representations from large-scale unlabeled images that transfer effectively to downstream tasks through linear probing or finetuning.

In medical imaging, early works explored domain-specific pretext tasks. Model Genesis~\cite{zhou2021models} learns representations by restoring images corrupted with transformations including non-linear intensity mapping, local pixel shuffling, and image in-painting. There are also works that have adapted natural image SSL methods with medical domain knowledge. For example, enhanced contrastive learning for medical images~\cite{tang2022self,chaitanya2023local,wu2024voco}; improving the 3D training recipe for masked autoencoder~\cite{wald2025revisiting}; applying DINO-based self-distillation to large-scale radiological datasets~\cite{perez2025exploring,xu2025generalizable}. These methods primarily focus on learning generic visual features through reconstruction or contrastive objectives. In contrast, MASS introduces a novel segmentation-based pretraining paradigm that explicitly learns both semantic understanding and spatial localization, directly aligning pretraining objectives with downstream medical image analysis tasks.

\noindent\textbf{Supervised Pretraining and Synthetic Data.}
Supervised pretraining on large-scale datasets, e.g. ImageNet~\cite{deng2009imagenet} has proven effective for transfer learning in natural and medical imaging~\cite{ridnik2021imagenet,sun2017revisiting,raghu2019transfusion}. In medical imaging, recent works pursue large-scale supervised segmentation pretraining: STU-Net~\cite{huang2023stu} leverages TotalSegmentator~\cite{wasserthal2023totalsegmentator} for whole-body CT pretraining, while SuPreM~\cite{li2025well} trains on 25 million annotated voxels from 9,000 abdomen CT scans covering 25 organs and 7 tumors. However, supervised pretraining in medical imaging faces fundamental limitations. First, annotations are confined to predefined taxonomies, limiting generalization to novel anatomical structures, pathological variants, or unseen diseases. Second, annotation requires substantial expertise, making it expensive to scale to larger datasets or new modalities. These constraints hinder general-purpose foundation models for diverse clinical applications. Synthetic data generation sidesteps annotation costs~\cite{tadokoro2024primitive}—Anatomix~\cite{dey2024learning} samples anatomical configurations from TotalSegmentator masks and renders synthetic CT images for pretraining. However, the distribution gap between synthetic and real images limits transfer performance. MASS addresses both challenges by using automatically generated class-agnostic masks for self-supervised learning on real images, achieving scalability without annotations while avoiding synthetic data's distribution gap.

\noindent\textbf{Universal Segmentation and Interactive Models.}
Universal medical image segmentation aims to learn across diverse tasks and modalities. Recent approaches include CLIP-Driven Universal Model~\cite{liu2023clip}, Hermes~\cite{gao2024training}. UniverSeg~\cite{butoi2023universeg}, and Iris~\cite{gao2025show} further introduce in-context learning for adaptability. These works demonstrate that task diversity improves generalization. However, these methods still require extensive pixel-level annotations during training, limiting scalability. SAM~\cite{kirillov2023segment,ravi2024sam} achieves strong interactive segmentation with exceptional zero-shot capability in finding boundaries on natural images through large-scale pretraining, inspiring medical adaptations~\cite{ma2024segment,cheng2023sammed2d,wang2310sam}. However, these medical SAM methods still require expert annotations for training and user interaction during inference. MASS differs fundamentally in three ways. First, it eliminates annotations through auto-generated class-agnostic masks, enabling annotation-free scalability. Second, while UniverSeg and Iris achieve diversity from multiple annotated datasets with predefined classes, MASS achieves open-set diversity from unlimited auto-generated masks across unrestricted anatomical and pathological structures. Third, unlike SAM-based methods that adapt SAM for prompted segmentation, MASS uses SAM only for mask generation as SAM's representations lack semantics~\cite{espinosa2024there}. MASS learns semantic understanding through mask-guided pretraining, transferring to segmentation, classification, and beyond.

\section{Method}
\label{sec:method}

We first motivate why mask supervision serves as an effective pretext task for medical imaging (\S\ref{method:mask_sup}), then describe how to generate masks automatically without annotations (\S\ref{method:mask_gen}), detail the mask-guided training procedure (\S\ref{method:mass_train}), and finally explain the rationale for how semantic understanding emerges without semantic labels.

\subsection{Mask Supervision}
\label{method:mask_sup}

At the core of our approach is learning representations from mask supervision, guided by the assumption that representations within each mask should be similar while representations across different masks should be distinct. Why masks? Semantic segmentation captures both ``what" structures exist and ``where" they are located, encoding global context and fine-grained spatial details essential for medical image understanding. This dual nature is clinically critical:  different sites of intracranial hemorrhage indicate distinct causes and treatment strategies, while tumors at different anatomical locations exhibit unique etiologies and biological behaviors. By jointly learning semantic categories and spatial localization, segmentation-based pretraining provides an effective foundation for diverse downstream tasks including both segmentation and classification. The challenge is annotation cost: pixel-wise delineation is time-consuming, requires expert knowledge, and is expensive, preventing segmentation from scaling as a pretraining method. In the following, we demonstrate that expert annotations are not necessary: masks automatically generated by existing tools suffice for effective representation learning.

\subsection{Annotation-Free Mask Generation}
\label{method:mask_gen}
Classical unsupervised segmentation techniques~\cite{kass1988snakes,shi2000normalized,achanta2012slic} partition images based on local similarity or boundary cues, while recent foundation models like SAM~\cite{kirillov2023segment,ravi2024sam} demonstrate remarkable zero-shot segmentation capabilities. These methods generate masks according to boundaries and local coherence but lack semantic labels—they capture ``where" but not ``what" objects are. MASS leverages such automatically generated class-agnostic masks for learning strong semantic representations, offering key advantages. First, these masks are open-set: unconstrained by predefined taxonomies, they naturally cover multiple granularities from organ-level structures and sub-anatomical regions to diverse pathologies, enabling generalization to unpredictable downstream tasks. Second, they are inherently scalable: generated automatically without human intervention, enabling application across large-scale datasets, modalities, and anatomical regions.

MASS is a general framework compatible with any mask generation method. We use SAM2~\cite{ravi2024sam} for its strong performance and ease of deployment, with comparisons to other methods in the experimental section. Critically, SAM2 is trained on natural images without medical knowledge, excelling at boundary detection. With appropriate preprocessing, it identifies any structure with clear boundaries, enabling unrestricted anatomical and pathological coverage. To generate 3D masks, we adapt SAM2 through a multi-step process, see Fig. \ref{fig:framework} (A): create 3-channel inputs using different intensity windows (CT) or quantile normalization ranges (MRI/PET) to capture tissue contrasts, uniformly sample 2D slices along the optimal imaging axis, apply SAM2's automatic mask generation with dense point prompts, and propagate masks through volumes using SAM2's video prediction capability, generating hundreds to thousands of 3D masks per volume. While SAM2 produces coarse and noisy masks due to the domain gap, its strong boundary localization makes these class-agnostic masks well-suited for representation learning across multiple scales without manual annotation. Details and examples of SAM2 generated masks are in Supp. \ref{mask_gen}.

\begin{algorithm}[t]
\caption{MASS implementation pseudo-code}
\label{alg:mass}
\footnotesize
\begin{algorithmic}[1]
\REQUIRE Unlabeled 3D images $\mathcal{X}$, auto-generated masks $\{\mathcal{M}_i\}$, encoder $E_\theta$, task encoding module $T_\phi$, decoder $D_\psi$
\ENSURE Trained parameters $\theta, \phi, \psi$
\FOR{epoch $= 1$ to $N_{\text{epochs}}$}
    \FOR{each batch}
        \STATE \textcolor{gray}{// Sample image and mask}
        \STATE $x \sim \mathcal{X}$ \hfill $\triangleright$ Sample unlabeled image
        \STATE $m \sim \mathcal{M}_x$ \hfill $\triangleright$ Sample auto-generated masks
        
        \STATE \textcolor{gray}{// Create augmented reference and query views}
        \STATE $x_s, y_s \leftarrow \text{Augment}(x, m)$ \hfill $\triangleright$ Reference view
        \STATE $x_q, y_q \leftarrow \text{Augment}(x, m)$ \hfill $\triangleright$ Query view
        
        \STATE \textcolor{gray}{// Task encoding from reference}
        \STATE $F_s \leftarrow E_\theta(x_s)$ \hfill $\triangleright$ Encode reference image
        \STATE $\mathcal{T} \leftarrow T_\phi(F_s, y_s)$ \hfill $\triangleright$ Extract task embedding
        
        \STATE \textcolor{gray}{// Query segmentation guided by task embedding}
        \STATE $F_q \leftarrow E_\theta(x_q)$ \hfill $\triangleright$ Encode query image
        \STATE $\hat{y}_q \leftarrow D_\psi(F_q, \mathcal{T})$ \hfill $\triangleright$ Decode with task guidance
        
        \STATE \textcolor{gray}{// Compute loss and update}
        \STATE $\mathcal{L}_{Seg} \leftarrow \mathcal{L}_{\text{Dice}}(\hat{y}_q, y_q) + \mathcal{L}_{\text{BCE}}(\hat{y}_q, y_q)$
        \STATE Update $\theta, \phi, \psi$ via backpropagation
    \ENDFOR
\ENDFOR

\end{algorithmic}
\end{algorithm}

\subsection{Mask-Guided Self-Supervised Learning}
\label{method:mass_train}

We adopt an in-context segmentation (ICS) framework for MASS using only automatically generated class-agnostic masks instead of expert annotations, see Figure~\ref{fig:framework} (B) and Algorithm~\ref{alg:mass}.

\noindent\textbf{In-context segmentation} is a learning paradigm where a reference image-mask pair $(x_s, y_s)$ defines the segmentation target, guiding segmentation of a query image $x_q$. Unlike traditional approaches that train separate models for predefined classes, in-context segmentation allows a single model $f_\theta$ to adapt to arbitrary tasks through conditioning on reference examples: $\hat{y}_q = f_\theta(x_q; x_s, y_s)$, where the reference implicitly defines ``what'' to segment without explicit semantic labels.

\noindent\textbf{MASS training procedure.} We follow the Iris~\cite{gao2025show} architecture with three components: image encoder $E_\theta$, task encoding module $T_\phi$, and mask decoder $D_\psi$. MASS is a flexible learning framework that works with arbitrary image encoders, while the decoder and task encoding module require slight customization to support in-context learning. For each iteration, we sample an unlabeled 3D image $x$ and its automatically generated masks $m \in \mathcal{M}$, then create two augmented views: reference $(x_s, y_s)$ and query $(x_q, y_q)$. Our augmentation employs spatial transformations (rotation, scaling, translation) applied to both images and masks to maintain correspondence, and appearance transformations (brightness, contrast, gamma, Gaussian noise) applied only to images. The reference mask $y_s$ provides location information (``where''), while appearance variations force the model to learn semantic consistency (``what'') across different visual representations.

The model encodes the reference image $F_s = E_\theta(x_s)$ and extracts task embeddings $\mathcal{T} = T_\phi(F_s, y_s)$ capturing what anatomical structure to segment. These embeddings guide the decoder to predict the query mask: $\hat{y}_q = D_\psi(E_\theta(x_q), \mathcal{T})$. We optimize $\{\theta, \phi, \psi\}$ jointly using $\mathcal{L}_{Seg} = \mathcal{L}_{\text{Dice}}(\hat{y}_q, y_q) + \mathcal{L}_{\text{BCE}}(\hat{y}_q, y_q)$.

\begin{table*}[t]
\centering
\footnotesize
\caption{Single-dataset segmentation performance (Dice, \%). All models are pretrained on each dataset's full training set. Results shown as mean (std) over 3 runs. Baseline methods require finetuning on few-shot samples, while MASS can perform in-context inference without finetuning, already surpassing baselines. Finetuning MASS further improves performance. The number in the brackets of ``Full supervision" indicates the number of labeled samples for full supervised training. Bold: best performance; underlined: second best.}
\label{tab:small_scale_ics}
\setlength{\tabcolsep}{0.45em}
\begin{tabular}{@{}lccc|ccc|ccc|ccc@{}}
\toprule
& \multicolumn{3}{c}{BCV} & \multicolumn{3}{c}{AMOS MR} & \multicolumn{3}{c}{SS H\&N} & \multicolumn{3}{c}{KiTS Tumor} \\
\cmidrule(lr){2-4} \cmidrule(lr){5-7} \cmidrule(lr){8-10} \cmidrule(lr){11-13}
Method & 1-shot & 5-shot & 10-shot & 1-shot & 5-shot & 10-shot & 1-shot & 5-shot & 10-shot & 5-shot & 30-shot & 60-shot \\
\midrule
Scratch            & 27.3(4.2) & 72.8(1.6) & 75.2(1.1) & 32.8(3.8) & 70.3(1.8) & 75.9(1.2) & 51.8(3.2) & 58.5(2.4) & 65.1(1.6)  & 5.7(2.8) & 35.7(3.2) & 45.9(2.4) \\
MG                 & 47.1(2.8) & 74.8(1.3) & 77.7(0.9) & 37.7(3.4) & 72.2(1.5) & \underline{78.5(1.0)} & 55.2(2.6) & 62.1(2.0) & 68.8(1.3) & 6.5(2.5) & 38.2(2.8) & 46.5(2.1) \\
MAE                & 46.3(3.0) & 74.2(1.4) & 78.0(1.0) & 36.2(3.5) & 73.8(1.4) & 77.4(1.1) & 54.2(2.7) & 59.5(2.2) & \underline{72.6(1.2)} & 6.8(2.4) & \underline{42.2(2.5)} & 47.8(2.0) \\
S3D                & 46.7(2.9) & 75.0(1.3) & 77.9(1.0) & 36.8(3.4) & \underline{73.9(1.4)} & 77.2(1.1) & 55.2(2.6) & 60.2(2.1) & 71.5(1.2) & 7.6(2.3) & 40.7(2.6) & 49.4(1.9) \\
SimCLR             & 44.9(3.1) & \underline{75.5(1.2)} & \underline{78.4(0.9)} & 35.9(3.6) & 71.9(1.6) & 78.0(1.0) & 53.6(2.8) & 59.4(2.2) & 65.8(1.5) & \underline{9.3(2.1)} & 41.5(2.4) & \underline{53.8(1.7)} \\
\rowcolor{lightgray} MASS-IC & \underline{65.5(1.8)} & 73.0(1.5) & 73.6(1.3) & \underline{62.1(2.1)} & 66.9(1.7) & 71.6(1.4) & \underline{59.3(1.9)} & \underline{62.6(1.6)} & 63.5(1.6) & 2.7(1.6) & 3.8(1.4)  & 4.7(1.2)  \\
\rowcolor{lightgray} MASS-FT & \textbf{68.8(1.5)} & \textbf{80.7(0.8)} & \textbf{83.7(0.6)} & \textbf{65.9(1.7)} & \textbf{79.3(0.9)} & \textbf{84.7(0.6)} & \textbf{66.9(1.4)} & \textbf{74.1(0.9)} & \textbf{78.2(0.7)} & \textbf{13.7(1.9)} & \textbf{64.3(1.5)} & \textbf{75.3(1.1)}\\ 
\midrule
Full supervision & \multicolumn{3}{c}{83.6 [23]} & \multicolumn{3}{c}{85.5 [38]} & \multicolumn{3}{c}{78.2 [39]} & \multicolumn{3}{c}{81.7 [159]} \\
\bottomrule
\end{tabular}
\vspace{-1.5em}
\end{table*}

\noindent\textbf{Learning semantics without semantic labels.}  MASS learns semantic identity implicitly through an invariance-based mechanism that requires no semantic supervision. The key lies in the in-context segmentation objective: given two augmented views $(x_s, y_s)$ and $(x_q, y_q)$ of the same image-mask pair, the model must predict that the same mask applies despite radical visual differences between views. Appearance augmentations change intensity distributions, alter texture patterns through blur and noise, and modify local contrast, yet the mask remains identical. Spatial augmentations change position, scale, and orientation, yet the mask still identifies the same structure. This creates an impossibility for shortcut learning: the model cannot rely solely on intensity matching (appearance variations break this), texture patterns (blur and noise destroy these), absolute position (spatial transformations eliminate this cue), or orientation-specific features (rotations invalidate these). The only solution is to learn what remains invariant across these transformations—the fundamental semantic identity of anatomical structures. The "what" emerges not from direct supervision but as the only explanation that reconciles why the same mask applies across wildly different visual appearances. Meanwhile, the "where" is learned through the decoder's need to produce spatially precise masks. Critically, training on thousands of open-set class-agnostic masks spanning multiple granularities, from whole organs to sub-anatomical regions to diverse pathologies, forces the model to learn broad medical concepts. Many of these are fundamental compositional primitives: texture patterns, boundary characteristics, spatial configurations, and intensity distributions. These building blocks transfer effectively to downstream tasks because even novel anatomical structures or pathologies share common visual primitives with the diverse masks encountered in pretraining.

\noindent\textbf{Downstream use.} MASS supports multiple deployment modes. \textit{Training-free in-context segmentation}: The pretrained model directly segments novel structures given one/few annotated examples as references without parameter updates. \textit{Task-specific finetuning}: MASS can be finetuned like standard segmentation models with fixed classes. \textit{Classification}: The encoder serves as a frozen feature extractor or can be fully finetuned.
\vspace{-0em}
\section{Experiments}
\label{sec:experiment}


We evaluate MASS through a comprehensive experimental study around three central questions: (1) Does mask-guided pretraining work for segmentation across data scales? We demonstrate effectiveness from small-scale single-dataset pretraining to large-scale multi-modal pretraining on 5K volumes. (2) Do the learned representations generalize beyond segmentation? We show strong transfer to classification tasks on entirely unseen pathologies using frozen encoders. (3) What design choices make MASS effective? We analyze the impact of data quantity and diversity, mask quality, mask generation methods, and architectures.

\noindent\textbf{Implementation.} We generate class-agnostic masks for all unlabeled images using SAM2 and store them for training. We use 3D ResUNet as the default backbone and follow the Iris~\cite{gao2025show} architecture for in-context learning. Additional implementation details are in Supp. \ref{implementation_details}.

\subsection{Segmentation Performance Across Scales}
We evaluate MASS's segmentation capabilities at two scales: small-scale pretraining on individual datasets to validate the core approach, and large-scale multi-modal pretraining to show scalability and cross-dataset transfer.

\begin{table*}[ht]
\centering
\footnotesize
\caption{Large-scale multi-modal pretraining segmentation performance (Dice, \%). Results shown as mean over 3 runs (standard deviations in supplementary). Numbers in brackets under ``Full supervision" indicate the number of labeled samples for full supervised training. Bold: best performance; underlined: second best.}

\label{tab:large_scale_ics}
\setlength{\tabcolsep}{0.5em}
\begin{tabular}{@{}lcc|cc|cc|cc|cc|cc|cc|cc|c@{}}
\toprule
& \multicolumn{2}{c}{BCV} & \multicolumn{2}{c}{AMOS MR} & \multicolumn{2}{c}{SS H\&N} & \multicolumn{2}{c}{KiTS Tumor} & \multicolumn{2}{c}{LiTS Tumor} & \multicolumn{2}{c}{AutoPET} & \multicolumn{2}{c}{BraTS T1CE} & \multicolumn{2}{c}{ACDC} & \multicolumn{1}{c}{Pelvic}  \\
\cmidrule(lr){2-3} \cmidrule(lr){4-5} \cmidrule(lr){6-7} \cmidrule(lr){8-9} \cmidrule(lr){10-11} \cmidrule(lr){12-13} \cmidrule(lr){14-15} \cmidrule(lr){16-17} \cmidrule(lr){18-18}
\# shot & 1 & 10 & 1 & 10 & 1 & 10 & 30 & 60 & 10 & 30 & 30 & 100 & 30 & 60 & 1 & 10 & 1 \\
\midrule

Full supervision & \multicolumn{2}{c|}{83.6 [23]} & \multicolumn{2}{c|}{85.5 [38]} & \multicolumn{2}{c|}{78.2 [39]} & \multicolumn{2}{c}{81.7 [159]} & \multicolumn{2}{c|}{63.2 [99]} & \multicolumn{2}{c|}{67.8 [983]} & \multicolumn{2}{c|}{72.8 [241]} & \multicolumn{2}{c|}{90.8 [70]} & 94.7 [80]\\
Scratch & 27.3 & 75.2 & 32.9 & 75.9 & 51.8 & 65.1 & 35.7 & 45.9 & 42.5 & 50.4 & 40.1 & 53.4 & 54.0 & 62.8 & 38.7 & 69.8 & 57.8 \\
\midrule
\textit{Supervised pretrain} \\
SuPreM & 63.9 & 83.6 & 55.1 & 82.1 & 66.1 & 75.6 & 64.1 & 78.1 & 53.9 & 62.7 & 48.8 & 64.8 & 60.3 & 70.8 & 55.9 & 82.3 & 85.4 \\
Iris (IC) & 83.2 & 85.4 & 83.5 & 86.4 & 78.4 & 80.1 & 78.2 & 80.2 & 59.2 & 63.3 & 65.2 & 69.5 & 48.6 & 79.8 & 86.5 & 88.2 & 69.0 \\
Iris (FT) & 83.4 & 85.5 & 83.6 & 86.3 & 78.5 & 80.3 & 78.3 & 80.3 & 59.4 & 64.6 & 67.2 & 70.2 & 60.7 & 71.9 & 86.9 & 90.1 & 86.9 \\\midrule
\textit{Self-supervised pretrain} \\
OM-MG & 49.0 & 78.4 & 38.8 & 78.6 & 61.3 & 68.8 & 41.1 & 51.7 & 48.1 & 52.2 & 44.6 & 59.7 & 58.5 & 69.6 & 46.8 & 74.7 & 76.7 \\
OM-MAE & 48.8 & 79.1 & 37.9 & 77.9 & 59.1 & \underline{73.0} & 47.4 & 52.4 & 40.5 & 52.2 & 43.4 & 58.8 & 56.3 & 70.0 & 45.4 & 73.8 & 72.2 \\
OM-S3D & 46.4 & 78.3 & 38.9 & 77.2 & 59.8 & 71.9 & 44.3 & 54.7 & 42.8 & 50.6 & 42.3 & 58.3 & \underline{59.4} & \underline{70.3} & 46.3 & 74.2 & 73.3 \\
OM-SimCLR & 45.6 & 80.2 & 37.0 & 78.2 & 58.8 & 67.5 & 46.8 & \underline{60.8} & 49.2 & \underline{55.8} & 45.4 & 60.2 & 56.0 & 68.6 & 48.9 & \underline{75.8} & 77.0 \\
AnatoMix & 53.1 & \underline{81.0} & 35.9 & \underline{78.8} & 48.3 & 66.7 & 40.6 & 44.1 & \underline{49.9} & 52.1 & \underline{46.1} & \underline{62.8} & 58.7 & 66.7 & 42.8 & 73.1 & 82.2 \\
Merlin & 50.1 & 78.0 & 37.9 & 78.3 & \underline{62.7} & 72.7 & \underline{51.1} & 58.0 & 49.2 & 55.1 & 41.8 & 56.3 & 53.2 & 61.2 & 45.8 & 74.9 & 79.3 \\
\rowcolor{lightgray} MASS (IC) & \underline{68.7} & 73.6 & \underline{66.0} & 71.6 & \underline{62.7} & 63.5 & 3.4 & 4.3 & 2.6 & 4.5 & 13.9 & 18.6 & 11.0 & 12.0 & \underline{69.8} & \underline{75.8} & \underline{89.9} \\
\rowcolor{lightgray} MASS (FT) & \textbf{70.2} & \textbf{84.2} & \textbf{74.3} & \textbf{85.0} & \textbf{70.0} & \textbf{78.9} & \textbf{68.5} & \textbf{79.1} & \textbf{56.1} & \textbf{64.5} & \textbf{50.2} & \textbf{65.2} & \textbf{63.0} & \textbf{72.3} & \textbf{75.7} & \textbf{90.0} & \textbf{92.8} \\
\bottomrule
\end{tabular}
\vspace{-1.0em}
\end{table*}

\subsubsection{Small-Scale Pretraining}

\textbf{Setup.} We evaluate MASS's representation learning through few-shot segmentation on individual datasets. Each dataset is split into train/validation/test sets. All methods pretrain on the same full training set (unlabeled for self-supervised methods), then finetune using only a few labeled examples and evaluate on the test set. This simulates practical scenarios where unlabeled data is abundant but expert annotations are expensive.

\noindent\textbf{Datasets.} We use four diverse datasets spanning different modalities, body regions, and clinical targets: BCV~\cite{bcv} (abdomen CT, 13 organs), AMOS MR~\cite{ji2022amos} (abdomen MRI, 13 organs), SS H\&N~\cite{structseg} (head-neck CT, 22 organs), and KiTS~\cite{heller2019kits19} (kidney tumor CT).

\noindent\textbf{Baselines.} We compare against standard self-supervised methods: random initialization (Scratch), Model Genesis (MG)~\cite{zhou2021models}, MAE~\cite{he2022masked}, S3D~\cite{wald2025revisiting}, and SimCLR~\cite{chen2020simple}. All baselines require finetuning for segmentation. MASS supports two modes: MASS-IC performs in-context inference without parameter updates, while MASS-FT finetunes on few-shot samples.

\noindent\textbf{Results.} Table~\ref{tab:small_scale_ics} presents few-shot segmentation performance, where all methods are pretrained on the same dataset split, thus isolating the pretraining objective. MASS demonstrates remarkable capabilities across all datasets. \textbf{Direct deployment}: MASS-IC achieves competitive performance without any finetuning, surpassing all baselines in 1-shot settings on BCV, AMOS MR, and SS H\&N, demonstrating that MASS learns semantically meaningful anatomical representations directly from pretraining. \textbf{State-of-the-art with finetuning}: MASS-FT achieves the best performance across all settings. On BCV 1-shot, MASS-FT outperforms the best baseline by 21.7 points. Remarkably, MASS-FT with only 10 shots (25-40\% of training data) matches full supervision performance on anatomies, demonstrating exceptional data efficiency.
\noindent\textbf{Anatomical vs. pathological structures.} Results reveal distinct patterns: on datasets with well-defined anatomies (BCV, AMOS MR, SS H\&N), MASS-IC performs strongly in training-free few-shot as these structures exhibit consistent appearance and spatial context. Conversely, on highly variable KiTS tumors, MASS-IC shows limited in-context performance (2.7\%), but MASS-FT (64.3\% at 30-shot) substantially outperforms baselines (best: 42.2\%), demonstrating that MASS's pretrained representations adapt effectively to pathological structures through finetuning. Detailed failure case analysis on pathology can be found in Supp. \ref{failure_case}.

\subsubsection{Large-Scale Multi-Modal Pretraining}

\textbf{Setup.} We evaluate MASS's scalability by pretraining on approximately 5K 3D volumes from 12 datasets spanning CT, MRI, and PET modalities across diverse anatomical regions. This heterogeneous corpus simulates real-world clinical repositories. We evaluate transfer through few-shot finetuning on both in-distribution datasets (overlapping with pretraining) and out-of-distribution datasets (entirely unseen during pretraining, BraTS T1CE, ACDC, Pelvic). Dataset details are in Supp. \ref{dataset_details}.

\noindent\textbf{Baselines.} We compare against SOTA large-scale pretrained models using their publicly released weights: OpenMind~\cite{wald2025openmind} self-supervised methods (MG, MAE, S3D, SimCLR on 114K multi-modal images), AnatoMix~\cite{dey2024learning} (synthetic data), Merlin~\cite{blankemeier2026merlin} (15K CT with language), SuPreM~\cite{li2025well} (supervised, 2.1K scans, 25 organs/7 tumors), and Iris~\cite{gao2025show} (supervised, 2K multi-modal images). Due to computational constraints, we use pretrained models rather than retraining all baselines on our pretraining corpus.  More details in Supp. \ref{baseline}.

\begin{table*}[htbp]
\centering
\scriptsize
\caption{Classification performance (AUC, \%) across different downstream datasets and training data percentages. Training sample sizes are shown in parentheses. Bold: best performance; underlined: second best.}
\label{tab:classification_results}
\resizebox{\textwidth}{!}{%
\begin{tabular}{lccc|ccc|ccc|ccc}
\toprule
 & \multicolumn{3}{c}{RSNA ICH} & \multicolumn{3}{c|}{Liver Trauma} & \multicolumn{3}{c|}{Kidney Trauma} & \multicolumn{3}{c}{Spleen Trauma} \\
\cmidrule(lr){2-4} \cmidrule(lr){5-7} \cmidrule(lr){8-10} \cmidrule(lr){11-13}
 & 5\% & 30\% & 100\% & 5\% & 30\% & 100\% & 5\% & 30\% & 100\% & 5\% & 30\% & 100\% \\
Method & (761) & (4566) & (15220) & (153) & (918) & (3061) & (152) & (913) & (3046) & (152) & (917) & (3057) \\
\midrule
\textit{Full training} \\
Scratch     & 72.8 & \underline{79.8} & \textbf{89.5} & 57.5 & \underline{74.4} & \underline{90.5} & 54.14 & \underline{75.0} & \textbf{90.0} & 62.5 & 82.08 & \underline{89.2} \\
\midrule
\textit{Frozen encoder} \\
OM-MG     & 57.1 & 61.7 & 70.0 & 52.1 & 57.8 & 61.6 & 53.2 & 57.9 & 58.8 & 50.6 & 52.2 & 60.5 \\
OM-SimCLR & 58.9 & 64.3 & 72.7 & 62.8 & 65.8 & 68.9 & \underline{56.7} & 58.5 & 60.3 & 69.1 & 71.3 & 73.5 \\
OM-MAE    & 53.0 & 56.4 & 63.9 & 57.6 & 58.9 & 61.2 & 48.9 & 50.7 & 52.4 & 49.7 & 53.3 & 60.0 \\
OM-S3D    & 56.7 & 59.4 & 64.9 & 58.8 & 60.8 & 62.4 & 52.9 & 53.7 & 56.3 & 53.8 & 55.5 & 60.3 \\
SuPrem          & \underline{73.5} & 75.4 & 78.3 & \underline{62.9} & 68.3 & 69.8 & 49.6 & 54.9 & 59.7 & \underline{73.5} & \underline{83.3} & 87.3 \\
AnatoMix        & 56.6 & 58.7 & 63.8 & 51.8 & 55.4 & 58.5 & 51.0 & 52.2 & 54.4 & 59.2 & 64.7 & 65.7 \\
Merlin          & 57.3 & 60.2 & 65.5 & 54.6 & 60.1 & 66.4 & 52.7 & 58.0 & \underline{65.4} & 62.8 & 67.0 & 70.3 \\
\rowcolor{lightgray} MASS   & \textbf{75.4} & \textbf{79.9} & \underline{81.5} & \textbf{70.5} & \textbf{86.7} & \textbf{92.3} & \textbf{57.2} & \textbf{82.9} & \textbf{90.0} & \textbf{76.0} & \textbf{85.5} & \textbf{90.6} \\
\bottomrule
\end{tabular}%
}
\vspace{-1.5em}
\end{table*}

\noindent\textbf{SSL Results.} Table~\ref{tab:large_scale_ics} demonstrates that large-scale multi-modal pretraining improves upon single-dataset results, validating MASS's scalability. \textbf{Substantial improvements over SSL baselines with less pretraining data}: MASS-IC achieves strong performance on anatomical structures in 1-shot settings (e.g. BCV, AMOS MR, Pelvic), demonstrating direct learning of consistent anatomical representations. With finetuning, MASS-FT reaches SOTA performance, substantially outperforming all self-supervised methods with improvements of +17.1 points (BCV 1-shot over AnatoMix), +36.4 (AMOS MR 1-shot), +7.3 (SS H\&N over Merlin), +17.4 (KiTS tumor 30-shot), and +10.6 (Pelvic 1-shot). \textbf{Cross-modal transfer}: MASS's modality-agnostic training enables robust performance across CT, MRI, and PET. On MRI tasks (AMOS MR, ACDC), MASS shows particularly strong results, demonstrating that annotation-free mask generation generalizes better than methods relying on single modality data (Merlin) or synthetic data (AnatoMix). \textbf{Data diversity and efficiency}: Methods pretrained on larger data (OpenMind: 114K images) show inconsistent improvements, suggesting that data diversity and pretraining objective matter more than raw quantity. MASS's strong performance with 5K volumes validates that mask-guided learning provides richer training signals than reconstruction or contrastive objectives.

\noindent\textbf{Comparison with supervised pretraining.} While direct comparison is inherently unfair due to MASS using zero annotations versus thousands of expert-labeled scans for supervised methods (Iris, SuPreM), we observe complementary strengths. Supervised methods (Iris, SuPreM) achieve superior performance on datasets overlapping with their pretraining annotations: Iris reaches 83.2\% on BCV 1-shot versus MASS 70.2\%, reflecting the value of task-specific expert supervision when evaluation targets align with pretraining taxonomies. However, on out-of-distribution datasets (BraTS, Pelvic), MASS demonstrates competitive or superior generalization: Pelvic 1-shot MASS-FT achieves 92.8\% versus Iris 86.9\% and SuPreM 85.4\%. The key trade-off is clear: supervised pretraining achieves 10-15 points higher performance when evaluation targets match annotation taxonomies, but requires thousands of expert-hours at prohibitive costs and struggles with cross-modal transfer and novel anatomies. MASS's annotation-free approach enables pretraining at larger scale, providing better scalability-generality trade-offs for foundation models serving diverse, evolving clinical applications with unknown future requirements.

\subsection{Generalization on Classification}
To evaluate whether MASS learns broadly transferable representations, we test frozen encoder performance on classification tasks with entirely unseen pathologies for all comparison methods.

\noindent\textbf{Setup.} We train only a classifier on top of fixed pretrained encoders on two large-scale, out-of-distribution datasets: RSNA ICH~\cite{flanders2020construction} (5-class intracranial hemorrhage, 15K scans) and RSNA Trauma Detection~\cite{hermans2024rsna} (severe trauma prediction for liver, kidney, spleen, 3K scans each). This protocol directly measures feature transferability without task-specific adaptation.

\noindent\textbf{Results.} Table~\ref{tab:classification_results} shows MASS achieves the best frozen encoder performance across nearly all settings. Reconstruction-based methods (MAE, S3D) show limited transfer (56-65\% AUC), while contrastive learning (SimCLR) performs better (72.7\%). Methods trained on limited distributions struggle: AnatoMix (synthetic data) reaches 63.8\% on RSNA ICH, and Merlin (abdominal CT) achieves 65.5\%. Supervised segmentation pretraining (SuPreM) consistently ranks second, validating that dense prediction tasks learn representations beneficial for classification.

\noindent\textbf{Frozen features vs. full training.} Most remarkably, MASS's frozen encoder matches or exceeds full supervised training from scratch with limited data. With 5\% RSNA ICH data (761 samples), MASS (75.4\%) surpasses full training from scratch (72.8\%). On trauma detection with 30\% data, MASS substantially outperforms full training: liver (86.7\% vs. 74.4\%), kidney (82.9\% vs. 75.0\%), spleen (85.5\% vs. 82.1\%). Only with large datasets (15K+ samples) does training from scratch become competitive. This demonstrates that mask-guided pretraining learns broadly transferable features beyond the segmentation pretext task.

\begin{table*}[t]
\centering
\scriptsize
\caption{Analysis experiments. (a) MASS learns strong representations from imperfect masks. (b) Impact of mask generation methods (c) MASS generalizes across different backbone architectures.}
\label{tab:analysis}

\begin{subtable}[t]{0.32\textwidth}
\centering
\caption{Learning from imperfect masks (1-shot ICS)}
\setlength{\tabcolsep}{0.4em}
\begin{tabular}{lccc|c}
\toprule
Dataset & Best & Avg & \%$>$40 & 1-shot \\
\midrule
BCV & 75.0 & 15.2 & 14\% & 65.5 \\
SS H\&N & 40.3 & 7.1 & 13\% & 59.3 \\
\midrule
\multicolumn{5}{l}{Best: highest Dice between any auto mask and GT} \\
\multicolumn{5}{l}{Avg: mean Dice for masks overlapping with GT} \\
\multicolumn{5}{l}{\%$>$40: percentage with GT overlap $>$ 40 Dice} \\
\bottomrule
\end{tabular}
\end{subtable}
\hfill
\begin{subtable}[t]{0.32\textwidth}
\centering
\caption{Mask generation methods (1-shot ICS)}
\setlength{\tabcolsep}{1em}
\begin{tabular}{lcc}
\toprule
Mask Source & BCV & SS H\&N \\
\midrule
TotalSegmentator & \textbf{80.7} & 13.5 \\
SAM2 & 65.5 & \textbf{59.3} \\
SLIC & 54.3 & 43.8 \\
\midrule
Full supervision & 83.6 & 78.2 \\
\bottomrule
\end{tabular}
\end{subtable}
\hfill
\begin{subtable}[t]{0.32\textwidth}
\centering
\caption{Backbone architectures}
\setlength{\tabcolsep}{1em}
\begin{tabular}{lcc}
\toprule
Backbone & 1-shot ICS & ICH Cls \\
 & (5 datasets) & (5\%) \\
\midrule
MASS-ResUNet & \textbf{73.87} & 75.42 \\
MASS-I3ResNet & 72.56 & \textbf{75.98} \\
\midrule
Difference & -1.31 & +0.56 \\
\bottomrule
\end{tabular}
\end{subtable}
\vspace{-1.5em}
\end{table*}

\begin{figure}
    \centering
    \includegraphics[width=0.95\linewidth]{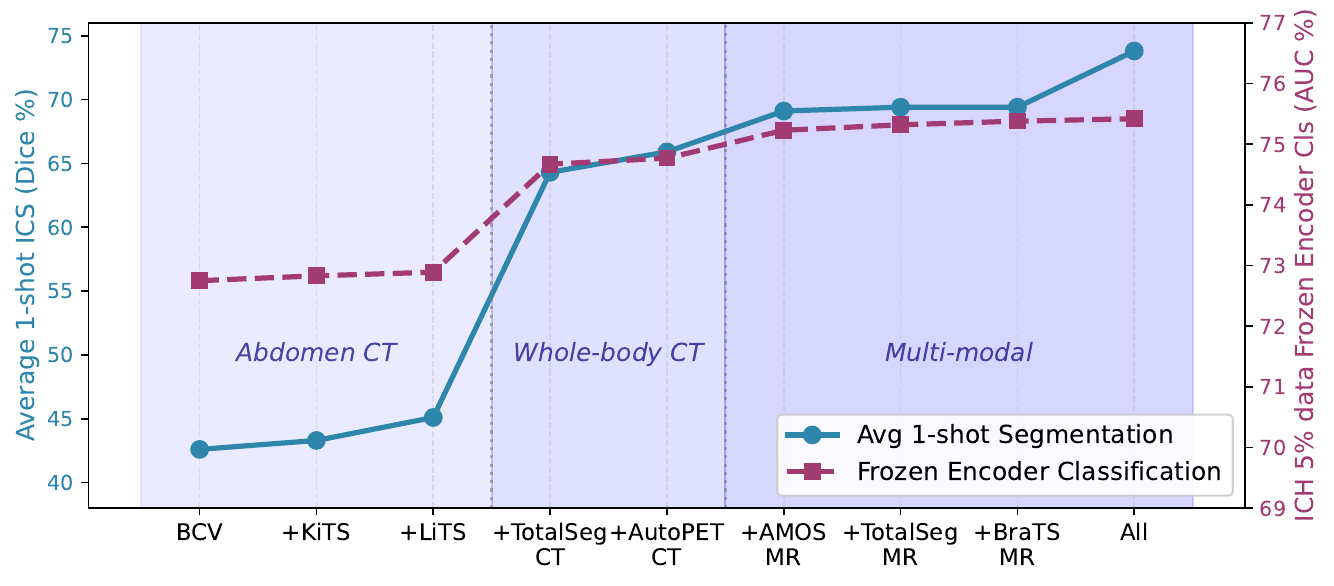}
    \caption{Progressive scaling analysis. Diversity along anatomical and modality dimensions drives generalization.}
    \label{fig:scale}
\vspace{-1.5em}
\end{figure}

\subsection{Analysis}

We analyze key factors contributing to MASS's effectiveness through systematic studies of data diversity, mask quality, and generation methods.

\noindent\textbf{Scaling with data diversity.} Figure~\ref{fig:scale} reveals that anatomical and modality diversity drive performance gains more than raw data quantity. \textit{Segmentation}: Starting from single-organ abdominal CT (BCV: 42.7\% average Dice across 5 datasets), adding similar abdominal CT datasets (KiTS, LiTS) yields minimal improvements. However, expanding to whole-body CT (TotalSeg) produces a dramatic jump, while subsequent multi-modal additions (AMOS MR, TotalSeg MR, BraTS) reach 73.9\%. \textit{Classification}: On out-of-distribution RSNA ICH (5\% data), pretraining on BCV alone (23 scans) achieves 72.8\% AUC—outperforming all SSL baselines and approaching SuPreM (73.5\%). Adding similar datasets provides negligible gains, while whole-body CT and multi-modal data drive substantial improvements. These results validate that MASS is highly data-efficient due to effective mask-guided supervision, and diversity along anatomical and modality dimensions drives continued improvement while incremental scaling within constrained domains saturates quickly.

\noindent\textbf{Learning from imperfect masks.} Table~\ref{tab:analysis} (a) quantifies automatically generated mask quality by measuring Dice overlap with ground truth annotations. On BCV, average overlap is only 15.2\% with just 14\% of masks achieving Dice $>$ 40; on SS H\&N, metrics drop to 7.1\% average. Despite this weak supervision, MASS achieves 65.5\% and 59.3\% one-shot ICS performance. This demonstrates that MASS learns semantic concepts beyond individual mask boundaries by training on diverse, approximately correct segmentation tasks, validating that weak, class-agnostic supervision suffices with appropriate learning frameworks.

\noindent\textbf{Mask generation methods.} Table~\ref{tab:analysis} (b) evaluates different ways to generate weak masks: TotalSegmentator~\cite{wasserthal2023totalsegmentator} (104 predefined anatomical classes), SAM2 (class-agnostic), and SLIC (superpixels). We use these methods to generate masks on BCV and SS H\&N and train with MASS. On BCV, TotalSegmentator achieves 80.7\% (approaching full supervision 83.8\%) as its taxonomy aligns with evaluation targets. However, on SS H\&N, TotalSegmentator collapses as its taxonomy excludes head-neck structures. In contrast, SAM2 maintains 59.3\%, demonstrating that class-agnostic masks enable adaptation to arbitrary anatomical structures. This suggests complementary use cases: predefined masks excel when targets align with existing taxonomies, while class-agnostic masks suit open-set unseen scenarios.

\noindent\textbf{Architecture generalization.} Table~\ref{tab:analysis} (c) validates MASS's architecture-agnostic nature by pretraining ResUNet (segmentation model) and I3DResNet152~\cite{carreira2017quo} (large encoder used by Merlin) on identical data and training protocol. Both achieve comparable performance with task-specific advantages: ResUNet excels at segmentation, while I3DResNet shows slight advantages for classification. These minimal differences demonstrate that pretraining benefits are independent of specific encoder design.

\section{Discussion and Conclusion}
\label{sec:discussion}

We present MASS, demonstrating that self-supervised pretraining with automatically generated masks eliminates expert annotation requirements while learning general-purpose medical imaging representations. The key insight is that semantic understanding emerges implicitly from invariance constraints—the model learns what fundamentally defines anatomical structures by finding what remains stable across radical appearance and spatial variations. By decoupling representation learning from expert annotation, MASS enables pretraining on arbitrarily large, diverse datasets with open-set masks that capture the full spectrum of anatomical and pathological diversity unconstrained by predefined taxonomies. This opens a practical path forward: rather than scaling annotation efforts, we can scale unlabeled data with weak supervision from automated tools, democratizing foundation model development for institutions and applications where expert annotation is prohibitively expensive or infeasible.

\noindent\textbf{Limitation and Future Work.} To fully demonstrate MASS as a robust self-supervised learning method, we deliberately exclude expert-annotated masks during pretraining. However, investigating the potential synergy between automatically generated weak masks and expert annotations during pretraining represents a promising avenue for future research. Additionally, scaling to larger datasets and incorporating higher-level self-supervised signals beyond segmentation could further enhance model performance. Finally, while we demonstrate that MASS successfully learns diverse medical concepts during pretraining, aligning these learned representations with textual modalities, such as radiology reports, remains an important direction for advancing vision-language understanding and automated report generation in medical imaging.

\noindent\textbf{Acknowledgment.} This work was supported in part by the Medical Imaging and Data Resource Center (MIDRC), which is funded by the National Institute of Biomedical Imaging and Bioengineering (NIBIB) under contract 75N92020C00021 and through the Advanced Research Projects Agency for Health (ARPA-H).
{
    \small
    \bibliographystyle{ieeenat_fullname}
    \bibliography{main}
}

\clearpage
\setcounter{page}{1}
\maketitlesupplementary







%

\section{Supplementary Method}

\subsection{Annotation-Free Mask Generation with SAM2}
\label{mask_gen}

\noindent\textbf{Pipeline Overview.} We adapt SAM2~\cite{ravi2024sam}, a foundation model trained on natural images, to generate class-agnostic 3D masks from medical volumes without any human annotation. Our pipeline consists of five stages: (1) \textit{Multi-channel preprocessing} converts single-channel medical volumes into 3-channel representations using complementary intensity mappings to preserve tissue contrasts compatible with SAM2's RGB input expectations, (2) \textit{Slice sampling with automatic axis selection} identifies the highest-resolution imaging axis and samples a sparse set of 2D slices based on physical spacing to ensure consistent coverage, (3) \textit{2D mask generation} applies SAM2's automatic segmentation mode with dense point prompts to systematically explore each slice, prioritizing diversity and coverage to generate hundreds of candidate masks per slice spanning multiple granularities, (4) \textit{3D propagation} extends 2D masks into volumetric segmentations using SAM2's video tracking capability, and (5) \textit{Post-processing} refines masks through connected component filtering, removes redundant overlapping masks. The entire pipeline operates at original image resolution, generating hundreds to thousands diverse masks per volume that serve as structural proposals for representation learning.

\noindent\textbf{Multi-Channel Preprocessing.} SAM2 expects 3-channel RGB images with pixel values in [0, 255], while medical images are single-channel grayscale volumes with modality-specific intensity ranges. For CT images with Hounsfield Units spanning -1024 to 3071, we create three channels using complementary intensity windows that capture different tissue contrasts. For abdomen CT, we apply: (1) soft tissue window (center=60 HU, width=350 HU), (2) contrast window (center=15 HU, width=250 HU), and (3) bone window (center=150 HU, width=1200 HU). For MRI and PET images, we apply quantile-based normalization with three percentile ranges: (5-95\%), (15-85\%), and (1-99\%). Each intensity mapping forms one channel of the final RGB representation.

We evaluated three preprocessing strategies (Table~\ref{tab:ablation_sam}): (1) \textit{Linear mapping}: map a fixed intensity range (e.g., [-400, 1500] HU for CT) to [0, 255] and replicate across three channels; (2) \textit{Single window}: apply one clinical window (e.g., soft tissue window) and replicate across three channels; (3) \textit{Three-channel windows}: create three distinct intensity mappings targeting different tissues. The three-channel approach (65.5\% Dice) substantially outperforms linear mapping (54.7\%) and single-window (64.5\%) strategies. This is because complementary tissue contrasts in each channel enable SAM2 to detect boundaries across diverse anatomical structures—tissues invisible in one window may have clear boundaries in another. We adopt three-channel windows as our default strategy, though users can tune specific window parameters to optimize SAM2's sensitivity for their target regions of interest.

\begin{table}[h]
\centering
\small
\caption{Ablation on SAM2 preprocessing strategies. Evaluated on BCV 1-shot segmentation.}
\label{tab:ablation_sam}
\begin{tabular}{lc}
\toprule
\textbf{Preprocessing Method} & \textbf{Dice (\%)} \\
\midrule
Linear mapping & 54.7 \\
Single window replication & 64.5 \\
Three-channel windows & \textbf{65.5} \\
Three-channel + multi-axis & 65.4 \\
\bottomrule
\end{tabular}
\end{table}

\noindent\textbf{Slice Sampling with Automatic Axis Selection.} Medical images typically exhibit anisotropic resolution with high in-plane resolution but coarse through-plane spacing. We automatically select the imaging axis with highest in-plane resolution for mask generation by computing average in-plane resolutions for each axis and selecting the minimum. For approximately isotropic images (spacing ratios $\leq$ 1.3), we default to the first axis. This optimizes mask quality while maintaining computational efficiency.

Along the selected axis, we sample 2D slices using physical distance intervals rather than fixed slice counts, ensuring consistent spatial coverage across images with different resolutions. Given axis spacing $s_{\text{axis}}$ and physical interval $d_{\text{mm}}$ (default: 15mm), the slice interval is $\max(1, \lfloor d_{\text{mm}} / s_{\text{axis}} \rfloor)$. This strategy balances comprehensive coverage with computational efficiency.

We also evaluate multi-axis processing, where masks are generated independently along all three anatomical axes (axial, sagittal, coronal) and merged (Table~\ref{tab:ablation_sam}). For the BCV dataset with anisotropic resolution (e.g. in-plane: 0.7×0.7mm, through-plane: 5-10mm), multi-axis processing yields 65.4\% performance—marginally lower than single-axis 65.5\% despite 3× computational cost. This occurs because processing lower-resolution planes (sagittal/coronal) introduces interpolation artifacts that produce degraded masks, which dilute rather than enhance the mask set quality. Based on these results, we default to single-axis processing on the highest-resolution plane for all experiments. However, for images with approximately isotropic resolution, where all axes have comparable quality, multi-axis processing may provide complementary structural information and improve performance.

\begin{figure*}
    \centering
    \includegraphics[width=1.0\linewidth]{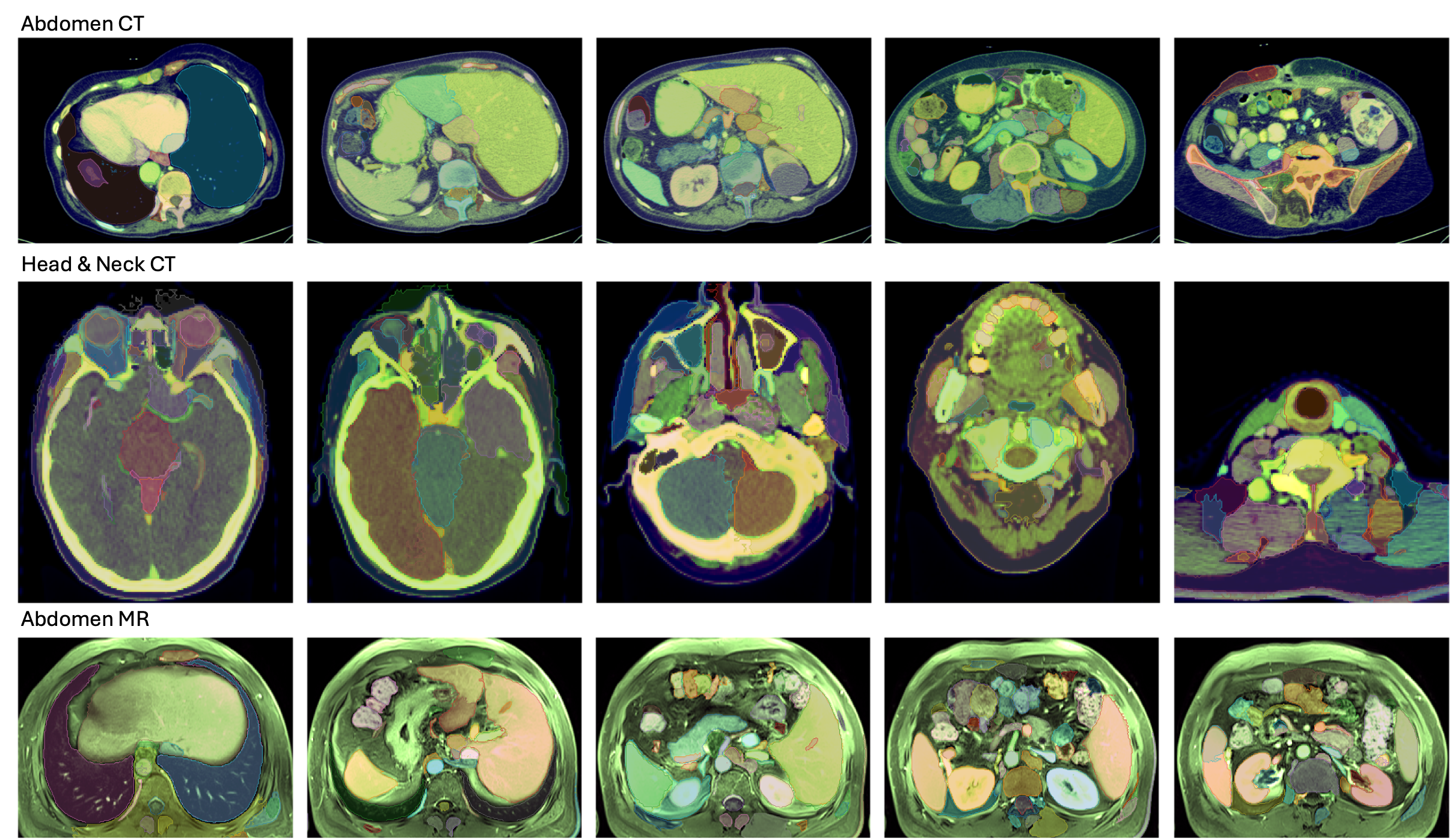}
    \caption{\textbf{SAM2 generated 2D masks on initial seed slices.} We show examples from abdomen CT, head \& neck CT, and abdomen MR. Images appear in color because we create 3-channel inputs for SAM2 using three complementary intensity windows. Each tissue's color reflects its relative intensity across the three channels. SAM2 generates meaningful region proposals with good coverage of diverse anatomical structures including organs, bones, muscles, sub-organ regions, and pathologies (cysts). However, the masks also contain substantial noise from over-segmentation, missing objects and struggle with small subtle structures. Despite these imperfections, the diverse mask proposals spanning multiple granularities provide sufficient supervision for MASS to learn generalizable representations.}
    \label{fig:sam_2d}
\end{figure*}

\noindent\textbf{2D Mask Generation.} For each sampled slice, we apply SAM2's automatic mask generation mode, which densely samples point prompts across the image grid and generates masks for each prompt without requiring any human annotation. This automatic mode systematically explores the entire image space to discover diverse segmentation proposals. We configure SAM2 with key parameters including points per side (typically 32), prediction IoU threshold (0.3-0.4), and stability score threshold (0.5-0.7) to control mask quality and quantity. Generated masks are filtered by minimum area to remove trivially small regions. We limit the number of masks retained per slice (typically 70) to control computation while maintaining diversity across granularities. Importantly, we prioritize diversity and coverage over individual mask accuracy—our goal is to generate masks spanning as many different anatomical and pathological structures and scales as possible, from large regions to fine-grained structures, rather than obtaining perfectly accurate segmentations. This stage generates hundreds to thousands of candidate masks per volume, providing rich structural proposals for representation learning.

\noindent\textbf{3D Propagation via Video Tracking.} To extend 2D masks into 3D volumes, we leverage SAM2's video prediction capability. We save the multi-channel slice sequence as JPEG frames and initialize SAM2's video predictor with this sequence. For each mask generated on a seed slice, we add it as a tracked object and perform bidirectional propagation: forward from the seed slice to the volume end, then backward from the seed slice to the volume start. This bidirectional approach ensures complete volumetric coverage while maintaining temporal consistency through SAM2's memory attention mechanism. The propagation produces per-object 3D mask volumes at the original image resolution without any resampling.

\noindent\textbf{Post-Processing} Each propagated 3D mask undergoes refinement: (1) \textit{Connected component filtering}: We identify all connected components and retain only the largest component that intersects with the original seed slice mask, removing spurious disconnected regions introduced during propagation. (2) \textit{Minimum volume filtering}: Masks smaller than a threshold are discarded to remove noise while preserving small but meaningful structures. (3) \textit{Redundancy reduction}: Since masks are propagated from different seed slices, the same anatomical structure may be captured multiple times with high overlap. To reduce redundancy while maintaining diversity, we identify pairs of masks with $>$90\% overlap (IoU $>$ 0.9) and randomly discard one from each pair.

\noindent\textbf{Mask Statistics} Our pipeline generates hundreds to thousands of masks per volume depending on anatomical complexity and dataset configuration. For the BCV dataset (abdominal CT scans), we generate an average of 1462 masks per scan, with minimum 938 and maximum 2210. 

\noindent\textbf{Computational Cost.} Mask generation is performed offline once per dataset. Typical processing time for a single CT volume (~512×512×150) with our default parameters: (1) Preprocessing: 10-20 seconds; (2) 2D mask generation (10-15 slices): 2-3 minutes; (3) 3D propagation: 3-5 minutes; (4)Post-processing: 30-60 seconds; Total: ~6-10 minutes per volume on a single NVIDIA H100 GPU. Once generated, masks are reused across all pretraining experiments, amortizing this one-time cost. The annotation-free mask generation pipeline can be parallelized across volumes, enabling efficient scaling to large datasets.

\begin{figure*}
    \centering
    \includegraphics[width=1.0\linewidth]{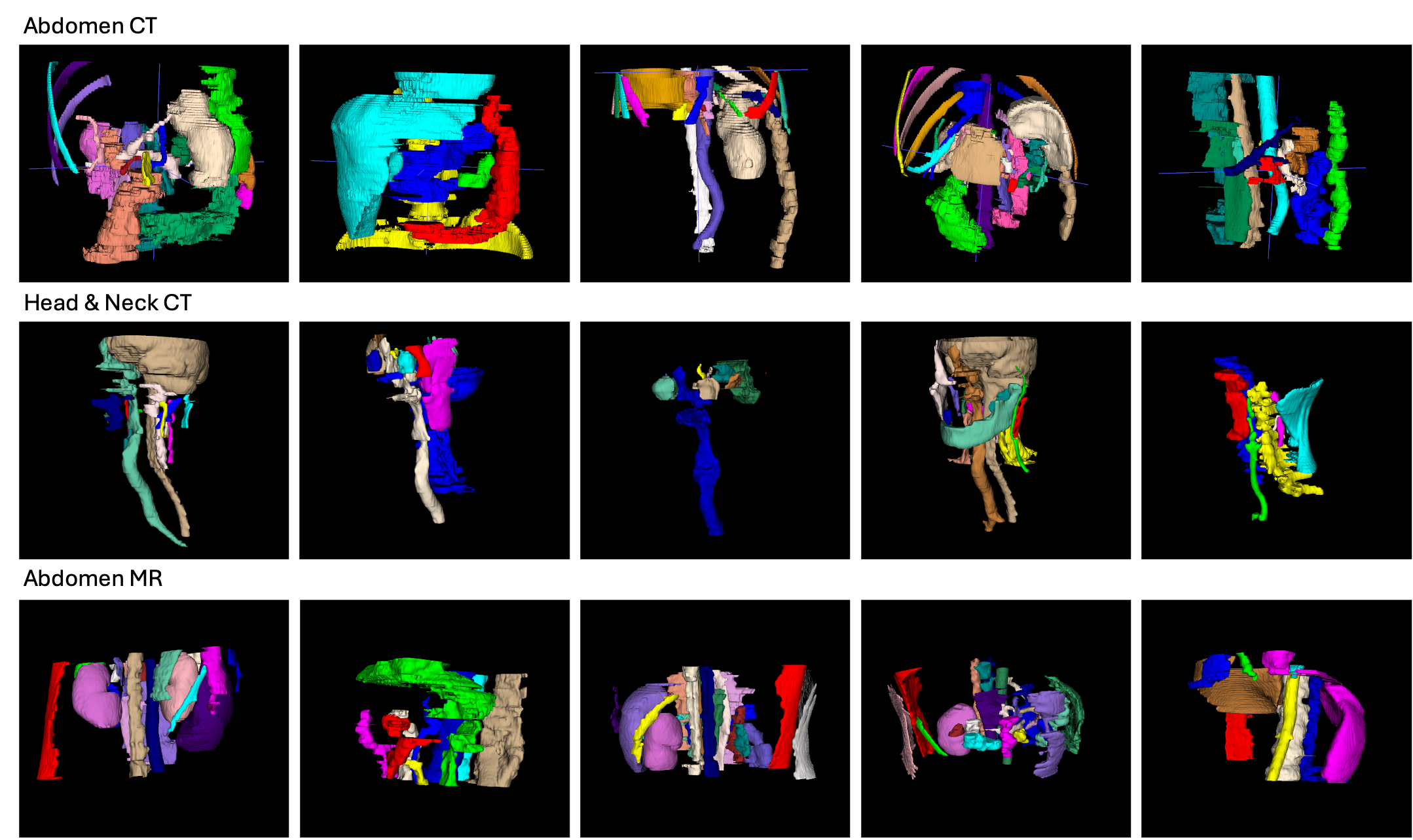}
    \caption{\textbf{SAM2 3D mask propagation results.} We show 3D masks generated by propagating the 2D seed masks from Figure~\ref{fig:sam_2d} through the volume using SAM2's video prediction capability. SAM2 successfully converts 2D masks into volumetric segmentations by tracking boundaries across slices. While propagation maintains anatomical coherence for most structures, the results still contain noise from inconsistent boundaries and occasional tracking failures. These imperfect but volumetrically consistent masks provide the structural supervision needed for MASS pretraining.}
    \label{fig:sam_3d}
\end{figure*}

\subsection{Visualization of SAM2 generated masks}

SAM2 enables fully automatic mask generation without any human annotation, providing the structural supervision needed for MASS pretraining. Figure~\ref{fig:sam_2d} shows 2D masks generated on seed slices for abdomen CT, head \& neck CT, and abdomen MR. The class-agnostic masks demonstrate excellent coverage and diversity, capturing anatomical structures at multiple granularities—from large organs (liver, kidneys) and skeletal structures (ribs, mandible, vertebrae) to finer sub-organ regions and small pathological findings (renal cysts). This diversity is critical for MASS: by training on masks spanning different anatomical scales and tissue types, the model learns compositional visual primitives applicable to novel structures. While the masks contain noise from over-segmentation and miss some small subtle structures (e.g., tiny vessels, thin organ boundaries), they provide sufficient structural information for learning generalizable representations.

Figure~\ref{fig:sam_3d} demonstrates the 3D propagation results using SAM2's video prediction capability. The same cases from Figure~\ref{fig:sam_2d} show how 2D seed masks are extended into volumetric segmentations by tracking boundaries through adjacent slices. Bidirectional propagation (forward and backward from each seed slice) ensures complete volumetric coverage. While propagation successfully maintains anatomical coherence for most structures, it introduces additional noise compared to the 2D masks. This occurs because SAM2's video tracking was trained on natural videos where occlusion handling is critical. When objects move behind others in natural scenes, the tracker maintains identity across the occlusion. However, in medical imaging, this behavior can inappropriately merge anatomically separate structures that happen to be adjacent. For example, in the second column of the first row in Figure~\ref{fig:sam_3d}, we observe the heart and liver connected as a single mask despite being distinct organs. Despite these imperfections, the propagated 3D masks provide volumetrically consistent structural supervision across hundreds to thousands of diverse proposals per volume, which proves sufficient for MASS to learn broadly transferable medical imaging representations.

\subsection{In-Context Segmentation Architecture}

\begin{figure}
    \centering
    \includegraphics[width=1\linewidth]{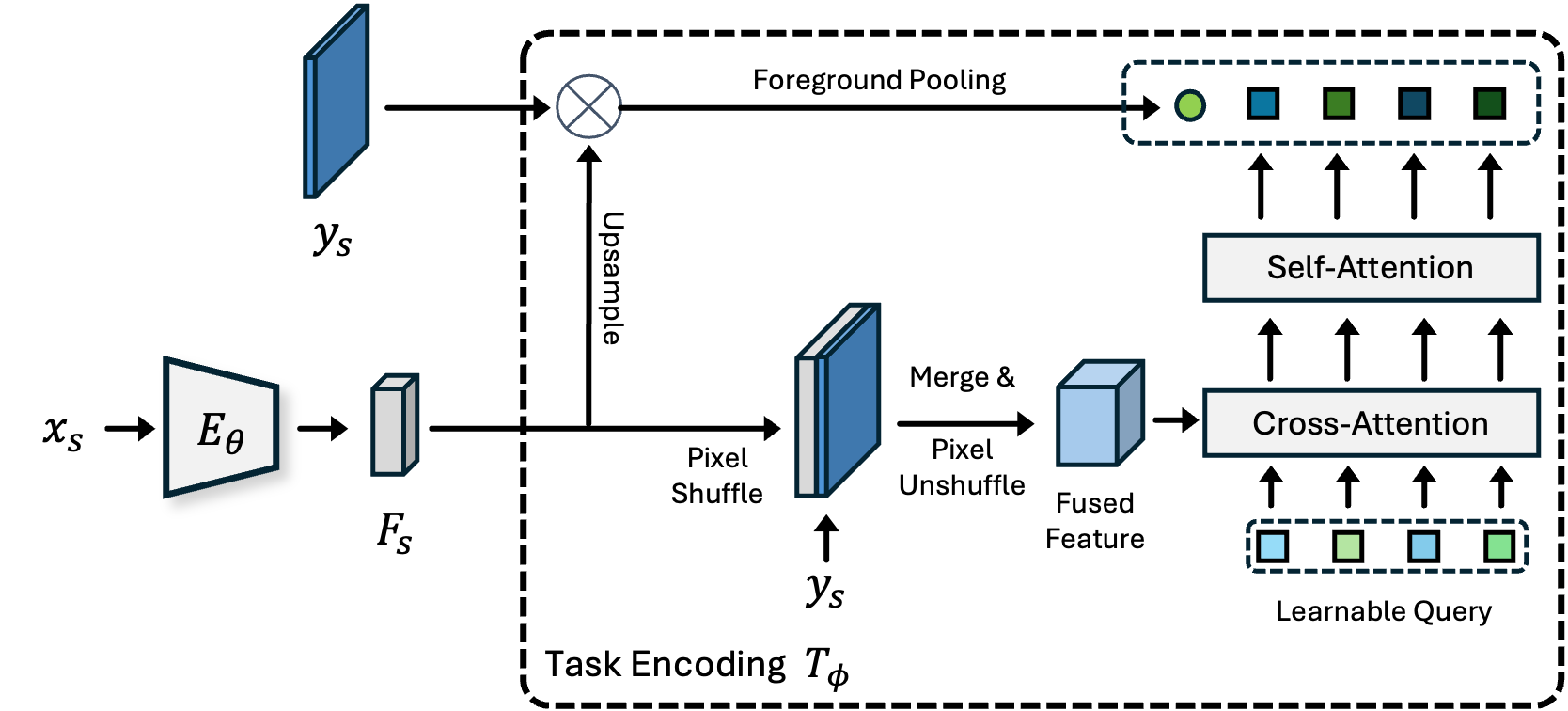}
    \caption{\textbf{Task encoding module architecture.} The module extracts compact task embeddings from reference image-mask pairs through two parallel streams: foreground feature encoding (top) captures fine anatomical details via high-resolution mask application, while contextual feature encoding (bottom) uses pixel shuffle operations and learnable query tokens with cross/self-attention to extract global context. The combined task embedding guides query image segmentation through the mask decoder. We follow~\cite{gao2025show} and refer the readers for more details in~\cite{gao2025show}}
    \label{fig:iris_arch}
\end{figure}

MASS adopts the Iris~\cite{gao2025show} architecture for in-context segmentation, which decouples task definition from query image inference through a lightweight task encoding module. As illustrated in Figure~\ref{fig:framework} (B), the framework consists of three components: an image encoder, a task encoding module, and a mask decoder. The details of the task encoding module is shown in Figure~\ref{fig:iris_arch}. Given a reference image-mask pair $(x_s, y_s)$, the encoder extracts visual features $F_s$, which the task encoding module then processes to produce compact task embeddings that capture "what to segment." The task encoding module operates through two parallel streams: (1) \textit{foreground feature encoding} upsamples encoder features to full resolution and applies the reference mask to preserve fine anatomical details, producing a pooled foreground embedding; (2) \textit{contextual feature encoding} employs pixel shuffle operations for memory-efficient high-resolution feature-mask fusion, followed by cross-attention and self-attention layers with learnable query tokens to extract contextual information. The final task embedding combines both foreground and contextual representations. During inference, the mask decoder takes query image features and task embeddings as inputs, using cross-attention to enable information exchange between task-specific guidance and query features, producing the final segmentation in a single forward pass.

Critically, MASS is a flexible learning framework compatible with arbitrary encoder decoder architectures. The image encoder can be any feature extraction network without modification, e.g. convolutional architectures (ResNet, UNet), vision transformers, or hybrid models. The decoder requires only minor customization: adding a few cross-attention layers to fuse query features with task embeddings. Beyond these cross-attention mechanisms for task conditioning, the decoder architecture itself is arbitrary. Users can employ UNet-style decoders, transformer decoders, or any other segmentation head. This architectural flexibility enables MASS to leverage advances in backbone design while maintaining its core self-supervised learning capability through mask-guided pretraining. We refer readers to the Iris paper~\cite{gao2025show} for comprehensive in-context segmentation architecture details and implementation specifics.

\begin{figure*}
    \centering
    \includegraphics[width=0.9\linewidth]{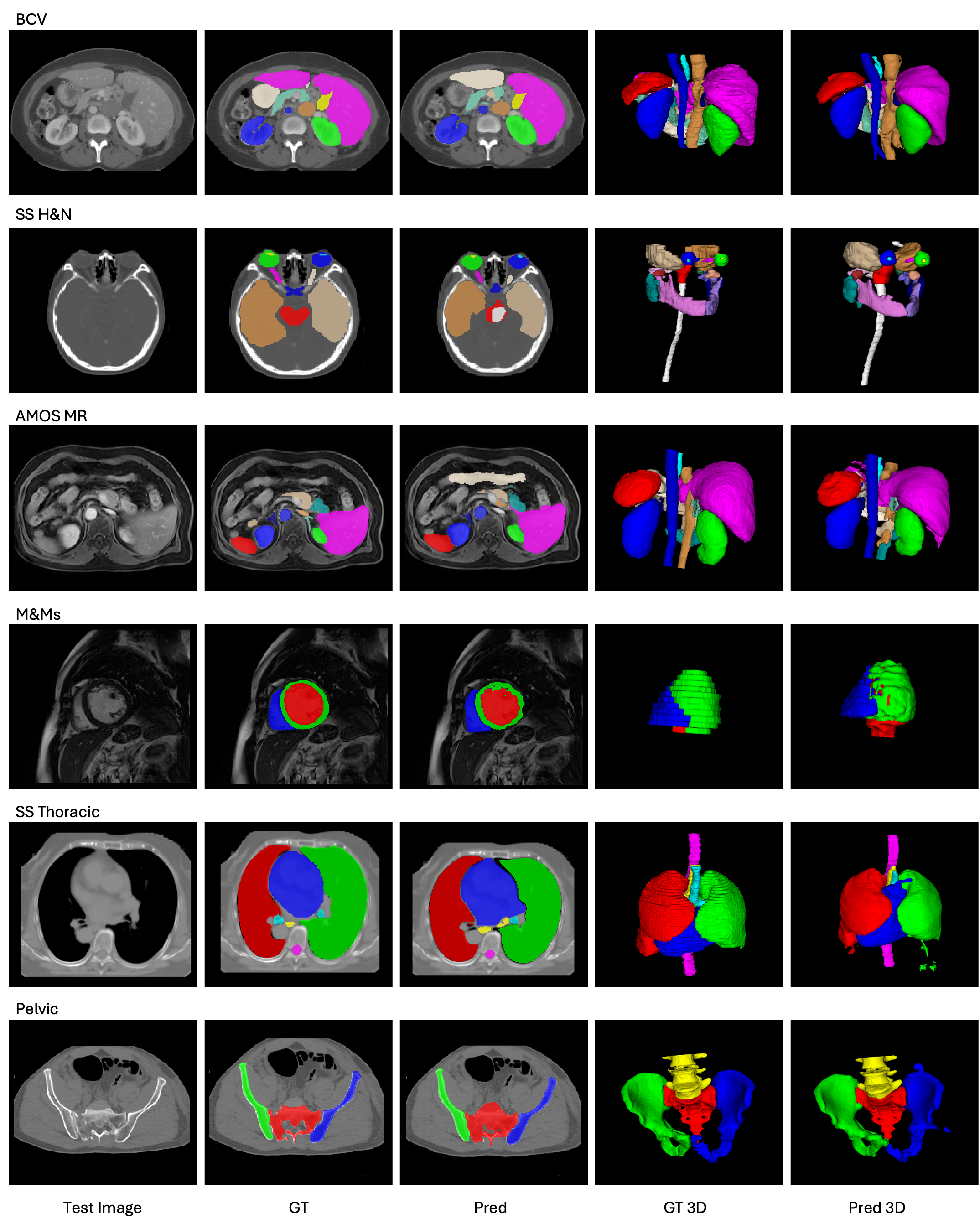}
    \caption{\textbf{Probing learned knowledge in pretraining through one-shot in-context inference.} We visualize the medical imaging understanding acquired by MASS during self-supervised pretraining by probing the model with one-shot in-context segmentation (no finetuning). Given a single reference image-mask pair, the pretrained model segments novel query images across diverse anatomical regions and modalities: BCV (abdominal CT), SS H\&N (head \& neck CT), AMOS MR (abdominal MRI), M\&Ms (cardiac MRI), SS Thoracic (thoracic CT), and Pelvic (pelvic CT). The results demonstrate that MASS learns rich medical imaging knowledge including organ morphology, spatial relationships, and anatomical consistency entirely from mask-guided self-supervision without expert annotations.}
    \label{fig:mass_direct}
\end{figure*}

\subsection{Implementation Details}
\label{implementation_details}

\noindent\textbf{Architecture.} For most of our experiments, we employ a 3D ResUNet encoder with four downsampling stages, producing feature maps at multiple resolutions. The task encoding module follows the Iris~\cite{gao2025show} architecture. The decoder consists of four upsampling stages with skip connections from the encoder, incorporating cross-attention layers at 3 stages (lower resolution) to fuse task embeddings with query features. All models are trained with randomly initialized weights unless otherwise specified.

\noindent\textbf{Training Configuration.} We train MASS for 100 epochs using LAMB optimizer with an initial learning rate of $2\times10^{-3}$, weight decay of $1\times10^{-5}$, and polynomial learning rate decay with power 0.9. Training employs mixed precision (BF16) on 8 NVIDIA H100 GPUs with PyTorch distributed data parallel (DDP). The effective batch size is 32 (4 per GPU). Loss combines Dice loss and binary cross-entropy with equal weighting. All 3D volumes are resampled to consistent spacing of 1.5mm.  The training 3D window size is cropped to $128\times128\times128$. Large-scale multi-modal pretraining (5K volumes) completes in 2 days.

\noindent\textbf{Finetuning Protocol.} For few-shot segmentation finetuning experiments, we fully finetune all parameters of all models. We train for 50 epochs using the same AdamW optimizer settings with learning rate $1\times10^{-4}$. Early stopping is applied based on validation performance. For in-context inference, we randomly sample $k$ reference images from the support (training) set and get the average task embedding then make predictions when $k>1$.

\noindent\textbf{Classification Protocol.} For frozen encoder classification, we attach classifier (attention pooling followed by a fully-connected layer) to the pretrained encoder. Only the classifier is trained using AdamW optimizer with learning rate $5\times10^{-4}$ for 50 epochs.

\section{Supplementary Experiments}

\subsection{Visualizing Learned Knowledge from Pretraining}

To demonstrate that MASS acquires rich medical imaging knowledge during self-supervised pretraining, we probe the pretrained model using one-shot in-context segmentation as a visualization tool. Figure~\ref{fig:mass_direct} shows qualitative results where the pretrained MASS model, given only a single reference example, segments novel anatomical structures across diverse body regions and modalities without any finetuning. This evaluation protocol serves as a direct probe into what anatomical concepts the model has internalized during pretraining.

The results reveal that MASS successfully learns fundamental medical imaging knowledge from mask-guided self-supervision: the model demonstrates understanding of organ morphology (shapes and sizes of anatomical structures), spatial relationships (relative positions of organs within the body), tissue boundaries (interfaces between different anatomical regions), and anatomical consistency (structures maintain coherent appearance across different patients). Critically, this knowledge emerges entirely from training on automatically generated class-agnostic masks without any expert annotations. The model has never seen expert-delineated boundaries or semantic labels during pretraining, yet it captures semantically meaningful anatomical concepts. The knowledge learned by MASS extends far beyond these shown examples—the visualization is inherently limited by our choice of reference prompts. The model has been exposed to thousands of diverse structures during pretraining through auto-generated masks, and we can only probe a small subset of this learned knowledge by selecting specific reference examples.

The visualization shows reasonable understanding across abdominal organs (BCV), head \& neck structures (SS H\&N), abdominal MRI (AMOS MR), cardiac structures (M\&Ms), thoracic organs (SS Thoracic), and pelvic bones (Pelvic). While the predictions are not perfectly aligned with expert annotation standards, this actually validates our pretraining paradigm: MASS is like the learning process of large language models, where broad knowledge is first acquired through self-supervision, then aligned with human expert standards through minimal supervised finetuning. Table~\ref{tab:large_scale_ics} quantifies this alignment process, demonstrating that just a few expert-annotated examples suffice to adapt the pretrained knowledge to downstream tasks with expert-level performance.

\begin{figure*}
    \centering
    \includegraphics[width=\linewidth]{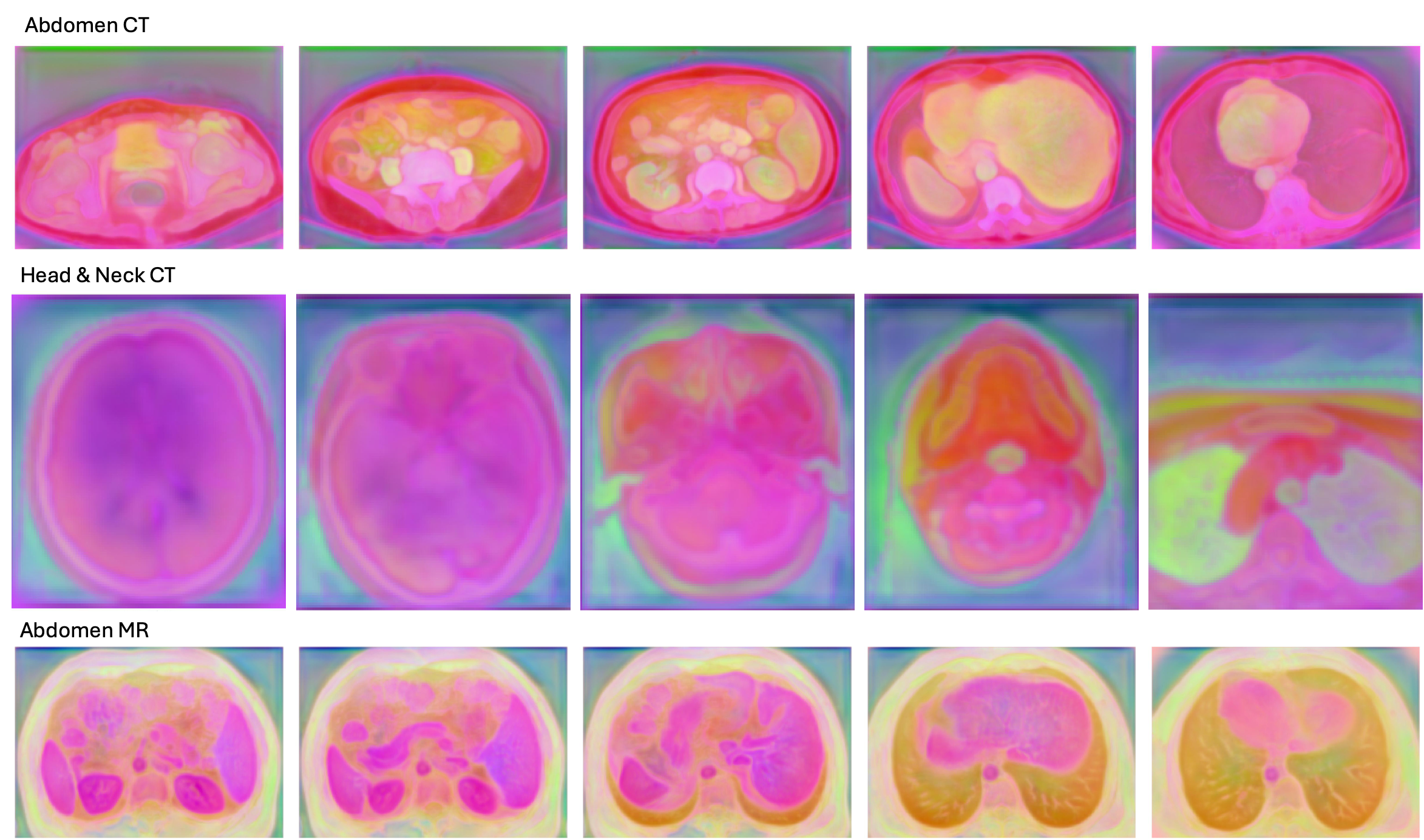}
    \caption{\textbf{PCA visualization of MASS learned features.} We extract decoder output features from the pretrained MASS model (without finetuning) and reduce to 3 channels via PCA for visualization. Examples shown include abdomen CT, head \& neck CT, and abdomen MR images. The visualizations reveal that MASS learns semantically meaningful representations with clear anatomical boundaries at full image resolution. Large structures (kidneys, liver, spleen, heart, lungs, bones, muscles) are clearly delineated, while semantically similar structures share consistent feature patterns (similar colors). Notably, MASS also captures fine-grained sub-anatomical details including pulmonary vasculature, hepatic vessels, and cardiac chambers, demonstrating multi-granular anatomical understanding learned entirely through mask-guided self-supervision without expert annotations.}
    \label{fig:pca}
\end{figure*}

\subsection{Feature Visualizations}

To qualitatively assess what representations MASS learns during self-supervised pretraining, we visualize the learned feature maps using Principal Component Analysis (PCA). Specifically, we use the large-scale pretrained MASS model as a feature extractor without any finetuning, extract the decoder output features, and reduce the channel dimension to 3 using PCA for RGB visualization. This approach reveals the semantic structure captured by MASS's learned representations.

Figure~\ref{fig:pca} presents PCA visualizations on representative examples from abdomen CT, head \& neck CT, and abdomen MR images. Several observations emerge from these visualizations:

\noindent\textbf{Clear anatomical boundaries.} Since MASS includes a decoder trained for segmentation, the output features maintain the original image resolution and exhibit remarkably sharp boundaries between anatomical structures. Organs such as kidneys, liver, spleen, heart, and lungs are clearly delineated, as are skeletal structures including pelvic bone, mandible, and skull. Soft tissue compartments (muscles, fat) also show distinct feature representations.

\noindent\textbf{Semantic consistency.} Semantically similar structures exhibit consistent feature representations (visualized as similar colors) across different spatial locations and patients. For instance, bilateral structures like left and right kidneys share similar feature patterns, indicating that MASS learns semantic concepts rather than merely spatial templates.

\noindent\textbf{Multi-granular understanding.} Most notably, MASS captures not only large anatomical structures but also fine-grained sub-anatomical details. The visualizations reveal clear representations of small structures such as pulmonary vasculature within the lungs, hepatic vessels within the liver, and distinct cardiac chambers. This multi-scale understanding emerges naturally from training on diverse auto-generated masks spanning multiple granularities.

These visualizations provide compelling evidence that mask-guided self-supervised pretraining enables MASS to learn rich, hierarchical medical imaging knowledge—from coarse organ-level semantics to fine sub-anatomical structures—entirely without expert annotations.

\subsection{Failure Case Analysis}
\label{failure_case}

Tables~\ref{tab:large_scale_ics} and~\ref{tab:small_scale_ics} reveal a notable performance gap: MASS's in-context segmentation (without finetuning) achieves strong results on anatomical structures but struggles significantly on pathologies such as tumors. We investigate this phenomenon to understand its underlying causes and implications for MASS's learned representations.

\begin{figure}
    \centering
    \includegraphics[width=\linewidth]{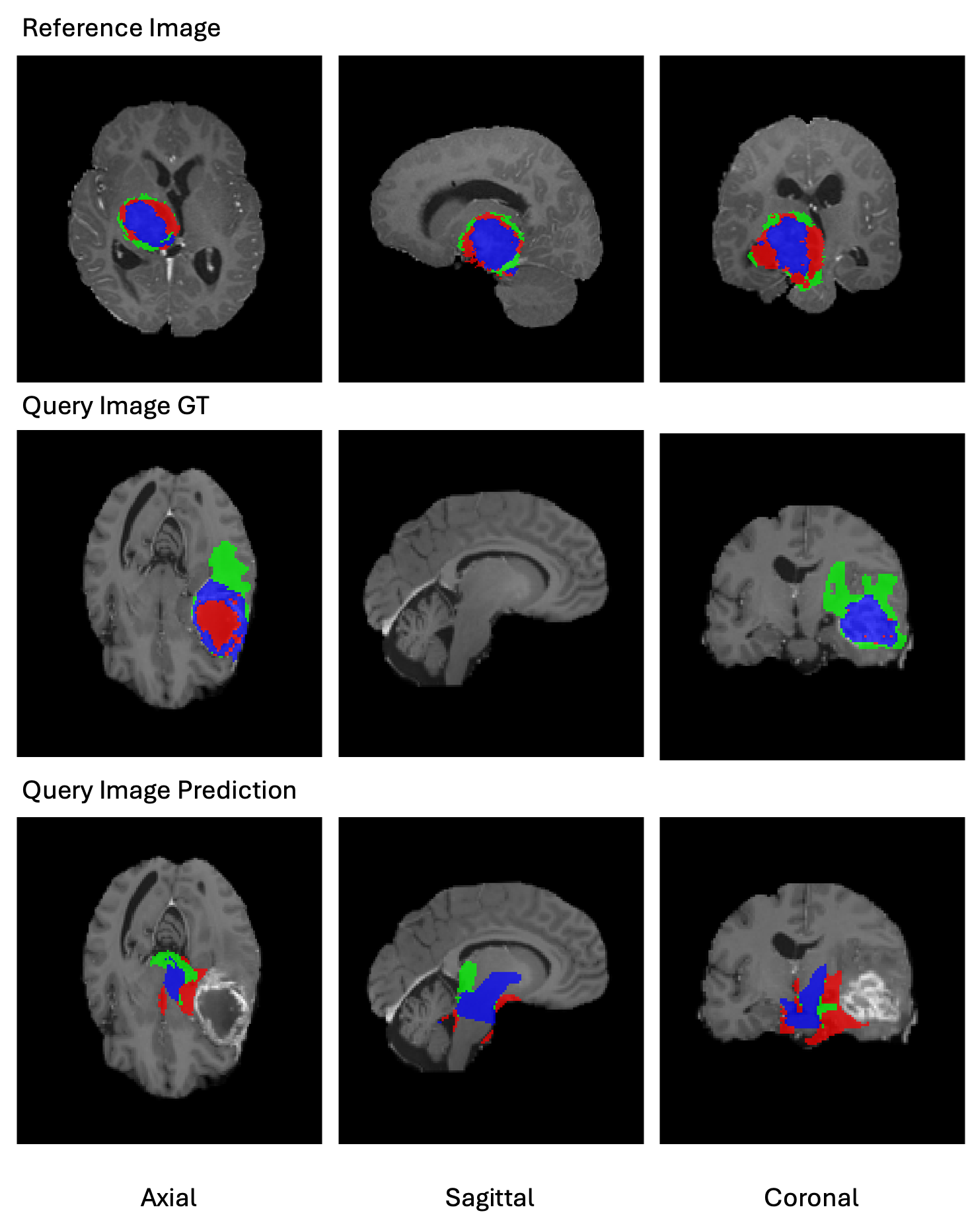}
    \caption{\textbf{Failure case analysis on pathology segmentation.} We visualize MASS's in-context segmentation on BraTS T1CE brain tumor data. Top row: reference image and mask defining the target tumor. Middle row: query image with ground truth tumor annotation. Bottom row: MASS's prediction, which incorrectly segments a region at a similar spatial location as the reference rather than the actual tumor. This failure stems from the large cross-patient variance in tumor location, size, and shape. MASS's pretraining objective learns within-image invariance but not cross-patient correspondence for highly variable pathologies. Notably, when using the query image as its own reference (with augmentation), MASS achieves $>$70\% Dice, confirming that the model learns meaningful tumor representations but lacks cross-patient semantic alignment. This gap is efficiently bridged through minimal finetuning with expert annotated samples.}
    \label{fig:failure_case}
\end{figure}

Figure~\ref{fig:failure_case} visualizes a representative failure case on BraTS T1CE brain tumor data. The first row shows the reference image and mask defining the segmentation target; the second row displays the query image with ground truth tumor annotation; the bottom row presents MASS's prediction. We observe that the model tends to segment regions in spatial locations similar to where the reference mask appears, rather than identifying the actual tumor in the query image. This behavior stems from the inherent characteristics of pathological structures: unlike anatomical organs that occupy consistent spatial positions across patients, tumors exhibit substantial variance in location, size, and shape. Consequently, MASS struggles to establish correspondence between the reference and query when the target structure appears in drastically different positions.

This limitation is actually expected given MASS's pretraining objective from two main factors. First, SAM2-generated masks for pathologies are inherently noisier due to their diffuse, irregular boundaries—unlike large organs with well-defined edges, tumors often exhibit gradual intensity transitions that challenge boundary-based segmentation. Second, MASS's pretraining objective learns invariance between two augmented views of the \textit{same} image, meaning the model is never exposed to how the same type of pathology appears across different patients. While anatomical structures maintain consistent spatial configurations across individuals (e.g., the liver is always in the right upper abdomen), pathologies can manifest anywhere within an organ with highly variable size, shape, and appearance. Thus, MASS learns strong within-patient invariance but lacks cross-patient correspondence for highly variable structures.

Crucially, this does not indicate that MASS fails to learn meaningful representations of pathologies. To verify this, we conducted an analysis experiment where we use the query image (and mask) itself as the reference (with different augmentations applied to reference and query views), then segment the query image. Under this self-reference setting, MASS achieves over 70\% Dice on BraTS T1CE, demonstrating that the model has indeed learned discriminative features for tumor tissues. The limitation lies not in representation quality but in cross-patient correspondence. MASS recognizes tumor features but does not inherently know that tumor features in different patients represent the same semantic concept.

This analysis explains why minimal finetuning with expert annotations dramatically improves pathology segmentation performance (Tables~\ref{tab:large_scale_ics} and~\ref{tab:small_scale_ics}). The few labeled examples teach MASS that diverse-appearing pathological features across patients belong to the same semantic category, effectively bridging the cross-patient correspondence gap. The strong finetuning results validate that MASS's pretrained representations capture meaningful pathological features, they simply require minimal supervision to align with expert-defined semantic categories. Improving such cross-patient correspondence would be an interesting future research direction.

\begin{table*}[ht]
\centering
\tiny
\caption{Large-scale multi-modal pretraining segmentation performance (Dice \%). Results shown as mean (standard deviation) over 3 runs. Numbers in brackets for "Full supervision" indicate the number of labeled samples for full supervised training. Bold: best performance; underlined: second best.}
\label{tab:large_scale_ics_std}
\setlength{\tabcolsep}{0.8em}
\begin{tabular}{@{}lcc|cc|cc|cc|cc|cc|cc|cc|c@{}}
\toprule
& \multicolumn{2}{c}{BCV} & \multicolumn{2}{c}{AMOS MR} & \multicolumn{2}{c}{SS H\&N} & \multicolumn{2}{c}{KiTS Tumor} & \multicolumn{2}{c}{LiTS Tumor} & \multicolumn{2}{c}{AutoPET} & \multicolumn{2}{c}{BraTS T1CE} & \multicolumn{2}{c}{ACDC} & \multicolumn{1}{c}{Pelvic}  \\
\cmidrule(lr){2-3} \cmidrule(lr){4-5} \cmidrule(lr){6-7} \cmidrule(lr){8-9} \cmidrule(lr){10-11} \cmidrule(lr){12-13} \cmidrule(lr){14-15} \cmidrule(lr){16-17} \cmidrule(lr){18-18}
\# shot & 1 & 10 & 1 & 10 & 1 & 10 & 30 & 60 & 10 & 30 & 30 & 100 & 30 & 60 & 1 & 10 & 1 \\
\midrule
Full supervision & \multicolumn{2}{c|}{83.6 [23]} & \multicolumn{2}{c|}{85.5 [38]} & \multicolumn{2}{c|}{78.2 [39]} & \multicolumn{2}{c|}{81.7 [159]} & \multicolumn{2}{c|}{63.2 [99]} & \multicolumn{2}{c|}{67.8 [983]} & \multicolumn{2}{c|}{72.8 [241]} & \multicolumn{2}{c|}{90.8 [70]} & 94.7 [80]\\
Scratch & 27.3(3.8) & 75.2(1.3) & 32.9(4.0) & 75.9(1.3) & 51.8(3.2) & 65.1(1.7) & 35.7(3.6) & 45.9(2.9) & 42.5(2.6) & 50.4(2.1) & 40.1(2.4) & 53.4(1.6) & 54.0(1.9) & 62.8(1.4) & 38.7(3.4) & 69.8(1.4) & 57.8(3.1) \\
\midrule
\textit{Supervised pretrain} \\
SuPreM & 63.9(1.9) & 83.6(0.8) & 55.1(2.1) & 82.1(0.9) & 66.1(1.7) & 75.6(1.0) & 64.1(1.2) & 78.1(0.9) & 53.9(1.6) & 62.7(1.2) & 48.8(1.5) & 64.8(0.9) & 60.3(1.2) & 70.8(0.9) & 55.9(2.0) & 82.3(0.8) & 85.4(1.3) \\
Iris (IC) & 83.2(1.2) & 85.4(0.7) & 83.5(1.1) & 86.4(0.7) & 78.4(1.0) & 80.1(0.7) & 78.2(0.9) & 80.2(0.7) & 59.2(1.3) & 63.3(1.0) & 65.2(1.1) & 69.5(0.8) & 48.6(1.6) & 79.8(0.8) & 86.5(1.1) & 88.2(0.7) & 69.0(1.9) \\
Iris (FT) & 83.4(1.0) & 85.5(0.6) & 83.6(0.9) & 86.3(0.6) & 78.5(0.8) & 80.3(0.6) & 78.3(0.8) & 80.3(0.6) & 59.4(1.2) & 64.6(0.9) & 67.2(1.0) & 70.2(0.7) & 60.7(1.0) & 71.9(0.8) & 86.9(0.9) & 90.1(0.6) & 86.9(1.1) \\\midrule

\textit{Self-supervised pretrain}\\
OM-MG & 49.0(2.6) & 78.4(1.2) & 38.8(2.9) & 78.6(1.2) & 61.3(2.3) & 68.8(1.4) & 41.1(1.9) & 51.7(1.5) & 48.1(1.9) & 52.2(1.4) & 44.6(1.7) & 59.7(1.1) & 58.5(1.4) & 69.6(1.0) & 46.8(2.5) & 74.7(1.2) & 76.7(2.1) \\
OM-MAE & 48.8(2.7) & 79.1(1.1) & 37.9(3.0) & 77.9(1.2) & 59.1(2.4) & 73.0(1.3) & 47.4(1.7) & 52.4(1.4) & 40.5(2.0) & 52.2(1.4) & 43.4(1.8) & 58.8(1.2) & 56.3(1.5) & 70.0(1.0) & 45.4(2.6) & 73.8(1.2) & 72.2(2.2) \\
OM-S3D & 46.4(2.8) & 78.3(1.2) & 38.9(2.9) & 77.2(1.2) & 59.8(2.4) & 71.9(1.3) & 44.3(1.8) & 54.7(1.4) & 42.8(1.9) & 50.6(1.5) & 42.3(1.8) & 58.3(1.2) & \underline{59.4(1.4)} & \underline{70.3(1.0)} & 46.3(2.6) & 74.2(1.2) & 73.3(2.1) \\
OM-SimCLR & 45.6(2.9) & 80.2(1.1) & 37.0(3.1) & 78.2(1.2) & 58.8(2.5) & 67.5(1.5) & 46.8(1.7) & \underline{60.8(1.3)} & 49.2(1.8) & \underline{55.8(1.3)} & 45.4(1.7) & 60.2(1.1) & 56.0(1.5) & 68.6(1.1) & 48.9(2.5) & \underline{75.8(1.2)} & 77.0(2.0) \\
AnatoMix & 53.1(2.4) & \underline{81.0(1.0)} & 35.9(3.2) & 78.8(1.2) & 48.3(2.8) & 66.7(1.5) & 40.6(2.0) & 44.1(1.7) & \underline{49.9(1.8)} & 52.1(1.4) & \underline{46.1(1.7)} & \underline{62.8(1.1)} & 58.7(1.4) & 66.7(1.1) & 42.8(2.7) & 73.1(1.3) & 82.2(1.8) \\
Merlin & 50.1(2.5) & 78.0(1.2) & 37.9(3.0) & 78.3(1.2) & \underline{62.7(2.2)} & 72.7(1.3) & \underline{51.1(1.6)} & 58.0(1.3) & 49.2(1.8) & 55.1(1.3) & 41.8(1.9) & 56.3(1.2) & 53.2(1.5) & 61.2(1.2) & 45.8(2.6) & 74.9(1.2) & 79.3(1.9) \\
\rowcolor{lightgray}\textbf{MASS (IC)} & \underline{68.7(1.7)} & 73.6(1.5) & \underline{66.0(1.9)} & 71.6(1.6) & \underline{62.7(1.8)} & 63.5(1.7) & 3.4(2.1) & 4.3(1.9) & 2.6(2.3) & 4.5(2.0) & 13.9(2.4) & 18.6(2.1) & 11.0(2.5) & 12.0(2.3) & \underline{69.8(1.6)} & \underline{75.8(1.4)} & \underline{89.9(1.4)} \\
\rowcolor{lightgray}\textbf{MASS (FT)} & \textbf{70.2(1.4)} & \textbf{84.2(0.8)} & \textbf{74.3(1.6)} & \textbf{85.0(0.7)} & \textbf{70.0(1.3)} & \textbf{78.9(0.9)} & \textbf{68.5(1.1)} & \textbf{79.1(0.8)} & \textbf{56.1(1.3)} & \textbf{64.5(1.0)} & \textbf{50.2(1.2)} & \textbf{65.2(0.8)} & \textbf{63.0(1.1)} & \textbf{72.3(0.9)} & \textbf{75.7(1.3)} & \textbf{90.0(0.7)} & \textbf{92.8(1.1)} \\
\bottomrule
\end{tabular}
\vspace{-0mm}
\end{table*}

\subsection{Additional Experiments}

\begin{table}[h]
\centering
\footnotesize
\caption{Comparison of training data of large-scale pretraining baseline methods.}
\label{tab:baseline_comparison}
\begin{tabular}{lllll}
\toprule
\textbf{Method} & \textbf{\# Samples} & \textbf{Modalities} & \textbf{Supervision} \\
\midrule
OpenMind & 114K & 24 MR & Self-supervised \\
Merlin & 15K & CT & Report \& EHR \\
AnatoMix & 120K & Synthetic & Synthetic labels \\
SuPreM & 2.1K & CT & 32 organ/tumor masks \\
Iris & 2K & CT, MR, PET & Expert annotated masks \\
MASS & 5K & CT, MR, PET & Self-supervised \\
\bottomrule
\end{tabular}
\end{table}

\noindent\textbf{Baseline selection rationale.}\label{baseline} Since MASS is a self-supervised learning method, we primarily compare against other self-supervised approaches: methods pretrained on OpenMind~\cite{wald2025openmind} (MG, MAE, S3D, SimCLR trained on 114K multi-modal images), AnatoMix~\cite{dey2024learning} (synthetic data generation), and Merlin~\cite{blankemeier2026merlin} (15K CT with language and EHR supervision). We also include two supervised pretraining methods—SuPreM~\cite{li2025well} (2.1K scans with 32 organ/tumor masks) and Iris~\cite{gao2025show} (2K multi-modal images)—to contextualize MASS's performance against methods that leverage expert annotations. Due to computational constraints, we evaluate all methods using their publicly released pretrained weights rather than retraining on our pretraining corpus. Table~\ref{tab:baseline_comparison} summarizes the pretraining data characteristics of each method.

We deliberately exclude SAM-based medical imaging variants~\cite{cheng2023sammed2d,ma2024segment} from our comparison for several reasons. First, these methods are finetuned in a supervised manner on very large-scale medical imaging datasets that have substantial overlap with our training and evaluation sets, making direct comparison unfair to MASS which uses no expert annotations. Second, we lack the computational resources to finetune these models on our data for a controlled comparison. Third, and most fundamentally, medical SAM methods are designed specifically for interactive segmentation with user prompts, whereas MASS aims at general-purpose representation learning that transfers beyond segmentation to tasks such as classification (Table~\ref{tab:classification_results}). The different design objectives make these methods not directly comparable.

Despite using substantially less data (5K volumes vs. 114K for OpenMind) and requiring zero expert annotations, MASS achieves superior performance across downstream tasks, demonstrating exceptional data efficiency and the effectiveness of mask-guided self-supervised learning. This comparison, while not perfectly controlled due to different pretraining corpora, reflects realistic scenarios where practitioners must choose between available pretrained models for their applications.

\noindent\textbf{Ablation on augmentation strategy.} Table~\ref{tab:ablation_aug} ablates augmentation design. Both spatial and appearance augmentations are necessary: spatial transformations maintain image-mask correspondence while appearance variations force semantic learning. Among magnitude levels, medium strength achieves optimal balance (65.5\%)—small augmentations provide insufficient variation while excessive augmentations introduce instability. The augmentation parameters we used for training MASS is: affine transformations with probability 0.8 including scaling ($\pm0.3$), rotation ($\pm30°$), and shear ($\pm0.1$); appearance transformations with probability 0.2 including brightness (multiplicative $[0.8, 1.3]$, additive std $0.15$), gamma ($[0.8, 1.3]$), contrast ($[0.8, 1.3]$), blur (sigma $[0.7, 1.5]$), and noise (std $0.04$). Spatial augmentations are applied jointly to images and masks to maintain correspondence; appearance augmentations are applied only to images.

\noindent\textbf{Inference Efficiency.} MASS is a pretraining framework that does not impose significant inference overhead compared to standard segmentation models. Table~\ref{tab:inference} reports inference benchmarks using the 3D ResUNet architecture on a single NVIDIA H100 GPU with BF16 precision. Full in-context inference (including reference encoding and query segmentation) takes 88.84ms per volume, achieving 11.26 FPS. Notably, the reference encoding step (39.72ms) only needs to be computed once per task—when segmenting multiple query volumes for the same target structure, only the query inference (49.00ms, 20.41 FPS) is required per volume. This makes MASS practical for batch processing scenarios. Peak memory usage is 6.3GB for input size $128^3$, comparable to standard 3D segmentation networks.

\noindent\textbf{Model Parameters.} Table~\ref{tab:inference} also presents the parameter breakdown. The full model contains 120.13M parameters, distributed across three components: the encoder (37.66M, 31.3\%), decoder (46.00M, 38.3\%), and task encoding module (36.35M, 30.3\%). The encoder and decoder together comprise a standard segmentation backbone (83.66M), while the task encoding module adds 36.35M parameters to enable in-context learning. This represents a moderate overhead ($\sim$43\%) compared to the base segmentation architecture. Importantly, the task encoding module is only used during reference processing; query inference primarily utilizes the encoder and decoder, resulting in efficient per-query computation. Furthermore, for downstream deployment with fixed target classes, the task encoding module can be entirely discarded. Only the encoder and decoder (83.66M) are required for standard segmentation finetuning. For classification tasks, only the encoder (37.66M) is needed as a feature extractor. This modular design allows practitioners to select the appropriate subset of components based on their deployment requirements. The overall model size remains comparable to typical 3D segmentation networks such as nnUNet and SwinUNETR, making MASS practical for clinical deployment.

\begin{table}[h]
\centering
\small
\caption{Model specifications and inference benchmark on NVIDIA H100 GPU with BF16 precision. Input size: $128^3$.}
\label{tab:inference}
\begin{tabular}{lc}
\toprule
\textbf{Metric} & \textbf{Value} \\
\midrule
\multicolumn{2}{l}{\textit{Inference Time}} \\
\quad Full in-context inference & 88.84 $\pm$ 0.13 ms \\
\quad Reference encoding & 39.72 $\pm$ 0.07 ms \\
\quad Query inference & 49.00 $\pm$ 0.02 ms \\
\quad Query inference FPS & 20.41 \\
\midrule
\multicolumn{2}{l}{\textit{Memory}} \\
\quad Peak memory & 6.3 GB \\
\midrule
\multicolumn{2}{l}{\textit{Parameters}} \\
\quad Encoder & 37.66M (31.3\%) \\
\quad Decoder & 46.00M (38.3\%) \\
\quad Task encoding module & 36.35M (30.3\%) \\
\quad Total & 120.13M \\
\bottomrule
\end{tabular}
\end{table}

\begin{table}[h]
\centering
\small
\caption{Ablation study on augmentation strategies. All experiments evaluated on BCV 1-shot segmentation.}
\label{tab:ablation_aug}
\begin{tabular}{lc|lc}
\toprule
\textbf{Aug. Strategy} & \textbf{Dice (\%)} & \textbf{Aug. Magnitude} & \textbf{Dice (\%)} \\
\midrule
Spatial only & 60.7 & Small & 61.3 \\
Appearance only & 54.6 & Medium & \textbf{65.5} \\
Both & \textbf{65.5} & Large & 62.9 \\
\bottomrule
\end{tabular}
\end{table}

\subsection{Dataset Details}
\label{dataset_details}

\begin{table}
  \caption{Dataset statistics. Upper section: upstream training datasets. Lower section: downstream evaluation datasets held out entirely from pretraining or reserved modalities.}
  \label{tab:dataset}
  \centering
   \scriptsize
   \setlength{\tabcolsep}{0.8mm}{
  \begin{tabular}{ccccccc}
    \toprule
    Dataset         & Body Region   & Modality  & Clinical Target   & \#Cls &   Size     \\
    \midrule
    AMOS CT~\cite{ji2022amos}  & Abdomen       & CT        & Organs            & 15        & 300   \\
    AMOS MR~\cite{ji2022amos}  & Abdomen       & MRI       & Organs            & 13        & 60    \\
    AutoPET~\cite{gatidis2022whole} & Whole body &  CT + PET & Lesions & 1 & 1014 \\
    BCV~\cite{bcv}     & Abdomen       & CT        & Organs            & 13        & 30    \\
    BraTS~\cite{menze2014multimodal} & Brain & T1/T2/FLAIR & Tumors & 3 & 213 \\
    KiTS~\cite{heller2019kits19}     & Abdomen       & CT        & Kidney \& Tumor    & 2         & 210   \\
    LiTS~\cite{bilic2019liver}     & Abdomen       & CT        & Liver \& Tumor    & 2         & 131   \\
    M\&Ms~\cite{campello2021multi} & Cardiac & cineMRI & Structures & 3 & 320 \\
    StructSeg H\&N~\cite{structseg} & Head \& Neck & CT   & Organs & 22 & 50\\
    StructSeg Tho~\cite{structseg}& Thorax        & CT        & Organs            & 6         & 50    \\
    TotalSeg CT~\cite{wasserthal2023totalsegmentator} & Whole body & CT & Anatomies & 104 & 1229\\
    TotalSeg MR~\cite{wasserthal2023totalsegmentator} & Whole body & MR & Anatomies & 50 & 299\\
    \midrule
    BraTS T1CE~\cite{menze2014multimodal} & Brain & T1CE MRI & Tumors & 3 & 213\\
    ACDC~\cite{bernard2018deep} & Cardiac & cineMRI & Structures &3 & 100\\
    Pelvic~\cite{liu2021deep}& Pelvic       & CT        & Bones    & 4         & 103 \\
    RSNA ICH~\cite{flanders2020construction} & Head & CT & Hemorrhage & 5 & 21744\\
    RSNA Trauma~\cite{hermans2024rsna} & Abdomen & CT & Trauma & 2 & 4710\\
    \bottomrule
  \end{tabular}}
\end{table}

This section provides comprehensive information about the datasets used in our experiments. We organize datasets into two categories: upstream training datasets used for pretraining MASS, and downstream evaluation datasets used to assess generalization capabilities. Table~\ref{tab:dataset} summarizes all datasets.

\noindent\textbf{Data Leakage Prevention.} We implement strict protocols to ensure no data leakage between pretraining and evaluation. For datasets used in both phases (BCV, AMOS, StructSeg, KiTS, LiTS, AutoPET), we partition data into non-overlapping train/validation/test splits at the patient level, using only training splits for pretraining and reserving test splits exclusively for evaluation. For BraTS, we additionally hold out the T1CE modality entirely from pretraining to evaluate cross-sequence generalization. Five datasets (ACDC, Pelvic, RSNA ICH, RSNA Trauma, and BraTS T1CE) are completely excluded from upstream training to serve as out-of-distribution evaluations. No test data from any dataset is used during MASS pretraining.

\subsubsection{Upstream Training Datasets}

\noindent\textbf{Multi-organ Abdominal Collection (AMOS).} AMOS~\cite{ji2022amos} is a multi-modal dataset featuring 500 CT and 100 MRI scans from 600 patients with abdominal abnormalities, acquired across eight scanner platforms. The dataset provides annotations for 15 anatomical structures in CT and 13 structures in MRI. We use both modalities in upstream training with the official training set (200 CT, 40 MRI), implementing a 95\%/5\% split for training/validation. The official validation set (100 CT, 20 MRI) is reserved exclusively for downstream evaluation.

\noindent\textbf{Whole-body PET/CT Collection (AutoPET).} AutoPET~\cite{gatidis2022whole} comprises 1,014 whole-body FDG-PET/CT studies, balanced between 501 cases with confirmed malignancies (lymphoma, melanoma, NSCLC) and 513 negative controls. We use both PET and CT modalities in upstream training with a 75\%/5\%/20\% patient-level split for training, validation, and testing.

\noindent\textbf{Abdominal CT from Multi-Atlas (BCV).} The BCV~\cite{bcv} collection consists of 30 abdominal CT scans with annotations for 13 abdominal organs. We implement a 75\%/5\%/20\% split (23/2/5 scans) for training, validation, and testing.

\noindent\textbf{Brain Tumor Segmentation (BraTS).} BraTS~\cite{menze2014multimodal} features multi-parametric MRI scans (T1, T1CE, T2, FLAIR) from 213 glioma patients with annotations for three tumor sub-regions. For upstream training, we use only T1, T2, and FLAIR modalities with a 75\%/5\%/20\% patient-level split. The T1CE modality is completely excluded from pretraining and reserved for downstream evaluation to test cross-sequence generalization.

\noindent\textbf{Kidney Tumor Dataset (KiTS).} KiTS19~\cite{heller2019kits19} comprises 210 contrast-enhanced CT scans from kidney cancer patients with annotations for kidney and tumor regions. We use a 75\%/5\%/20\% split for training, validation, and testing.

\noindent\textbf{Liver Tumor Segmentation (LiTS).} LiTS~\cite{bilic2019liver} contains 131 abdominal CT scans with annotations for liver parenchyma and tumor lesions. We employ a 75\%/5\%/20\% split for training, validation, and testing.

\noindent\textbf{Multi-Centre Cardiac Segmentation (M\&Ms).} M\&Ms~\cite{campello2021multi} features 320 cardiac cine-MRI scans from multiple centers and vendors with annotations for left ventricle, right ventricle, and myocardium. We use a 95\%/5\% split for training and validation. This dataset is used only for upstream pretraining; ACDC serves as the held-out cardiac evaluation dataset.

\noindent\textbf{Radiation Treatment Planning (StructSeg).} StructSeg~\cite{structseg} comprises CT imaging for radiation therapy planning. StructSeg H\&N includes 50 scans with annotations for 22 head \& neck organs-at-risk. StructSeg Tho contains 50 scans with annotations for 6 thoracic organs. We implement a 75\%/5\%/20\% split for both components.

\noindent\textbf{TotalSegmentator.} TotalSegmentator~\cite{wasserthal2023totalsegmentator} provides whole-body segmentation datasets: 1,229 CT scans with 104 anatomical structures and 298 MR scans with 50 structures. We use an 80\%/10\%/10\% split for training, validation, and testing. TotalSegmentator is only used for upstream pretraining due to its great anatomy coverage.

\noindent\textbf{Upstream Training Corpus Summary.} For large-scale multi-modal pretraining, we combine training splits from all upstream datasets, totaling approximately 5K 3D volumes spanning CT, MRI, and PET modalities. This corpus covers diverse anatomical regions (whole-body, abdomen, cardiac, brain, head \& neck, thorax) and clinical targets (organs, tumors, lesions), simulating real-world clinical repositories.

\subsubsection{Downstream Evaluation Datasets}

The following datasets are held out entirely from upstream training or represent reserved modalities, used exclusively to evaluate out-of-distribution generalization.

\noindent\textbf{Brain Tumor Segmentation - T1CE (BraTS T1CE).} We use the T1CE modality from BraTS~\cite{menze2014multimodal}, which was completely excluded from upstream training (only T1, T2, FLAIR were used). This evaluates cross-sequence generalization on the same patients but an unseen MRI contrast. We use the same 75\%/5\%/20\% patient-level split as upstream BraTS.

\noindent\textbf{Automated Cardiac Diagnosis Challenge (ACDC).} ACDC~\cite{bernard2018deep} consists of 100 cardiac cine-MRI scans from a different institution (University Hospital of Dijon) than M\&Ms, with different scanner configurations (Siemens 1.5T/3.0T). This dataset is entirely excluded from upstream training to serve as an out-of-distribution cardiac evaluation. We use a 75\%/5\%/20\% split.

\noindent\textbf{Pelvic Bone Segmentation (Pelvic).} Pelvic1K~\cite{liu2021deep} contains 103 CT scans annotated for four pelvic skeletal structures. This dataset is entirely excluded from upstream training to test generalization to novel anatomical regions not seen during pretraining. We employ a 75\%/5\%/20\% split.

\noindent\textbf{RSNA Intracranial Hemorrhage Detection (RSNA ICH).} RSNA ICH~\cite{flanders2020construction} comprises 21,744 non-contrast head CT scans with multi-label annotations for five hemorrhage subtypes. This dataset is entirely excluded from upstream training and evaluates transfer to pathology classification, a different task (classification vs. segmentation) on an unseen pathology. We split into 15,220/2,174/4,350 for train/validation/test, evaluating frozen encoder performance with 5\%/30\%/100\% of training data tuning.

\noindent\textbf{RSNA Abdominal Trauma Detection (RSNA Trauma).} RSNA Trauma~\cite{hermans2024rsna} contains 4,710 contrast-enhanced abdominal CT scans with annotations for trauma injury in abdomen organs. We use the severe trauma on three solid organs, liver, kidney, and spleen, for evaluation. This dataset is entirely excluded from upstream training. Although abdominal CT appears in pretraining, trauma detection represents a novel clinical task distinct from anatomical or tumor segmentation. We split into 65\%/5\%/30\% for train/validation/test, evaluating frozen encoder performance with 5\%/30\%/100\% of training data across three organ-specific binary classification tasks.

\end{document}